\documentclass{article}

\usepackage[final,nonatbib]{ml4eng_2020}

\usepackage[utf8]{inputenc} 
\usepackage[T1]{fontenc}    
\usepackage{hyperref}       
\usepackage{url}            
\usepackage{booktabs}       
\usepackage{amsfonts}       
\usepackage{nicefrac}       
\usepackage{microtype}      
\usepackage{amsmath}
\usepackage{xcolor}
\usepackage{hyperref}
\usepackage{xcolor}
\usepackage{caption}
\usepackage{subcaption}
\usepackage{svg}
\usepackage[warn]{textcomp}  
\usepackage{xspace}
\usepackage{graphicx}

\newcommand{\alg}{{DGMLG}\xspace}
\DeclareMathOperator{\Tr}{Tr}

\usepackage[
backend=biber,
sorting=none,
maxcitenames=2,  
]{biblatex}
\title{Building LEGO\\Using Deep Generative Models of Graphs}

\author{%
  Rylee Thompson\\
  University of Guelph\\
  \texttt{rylee@uoguelph.ca} \\
   \And
   Elahe Ghalebi, Terrance DeVries and Graham W.~Taylor\\
   University of Guelph \\
   Vector Institute \\
   \texttt{\{eghalebi,terrance,gwtaylor@uoguelph.ca\}}
}

\begin{document}

\maketitle

\begin{abstract}
Generative models are now used to create a variety of high-quality digital artifacts. Yet their use in designing physical objects has received far less attention. In this paper, we advocate for the construction toy, LEGO, as a platform for developing generative models of sequential assembly. We develop a generative model based on graph-structured neural networks that can learn from human-built structures and produce visually compelling designs. Our code is released at: \textcolor{blue}{\url{https://github.com/uoguelph-mlrg/GenerativeLEGO}}.
\end{abstract}

\section{Introduction}

Sequential assembly is the process of creating a desired form by connecting a series of geometric primitives. For example, furniture may be constructed from wooden segments, walls from individual bricks, and quilts from fabric patches.
When the individual pieces are few and simple with a limited number of interlocking structures, this facilitates assembly by humans or robots \cite{Zhang2020-pd}. Applications of automated assembly include modular packaging, or at a larger scale, pre-fabricated buildings.

Generative models can now create high-quality and diverse digital artifacts, particularly in the domain of images~\cite{karras2019style} and text~\cite{brown2020language}. Generative models are also being applied to physical design, such as computational chemistry and materials but they are less prevalent in architecture, landscape design, or manufacturing. Motivated by the opportunity of generative design of our physical environment, in this paper we explore the use of generative models to modular physical design.

Advances in generative models in areas such as images and text is in part due to the availability of data and ease of experimentation. Other areas of machine learning, such as reinforcement learning, have benefited from the availability of simulators. Simulators are conducive to experimentation supporting algorithm development and, at the same time, entertaining and familiar (c.f.~the Arcade Learning Environment \cite{Bellemare2013-js}). In the context of physical design, specifically the sequential assembly of modular structures, we believe that there is an analogue: the plastic construction toy, LEGO. Structures built from LEGO have a number of interesting properties: (i) they are complex enough to approach real-world design; ii) they are familiar and fun enough to attract interest from the generative models community; and iii) there is a huge amount of official and user-created structures available as a source of training data.

Learning to sequentially assemble LEGO structures based on human-generated examples is a sequential decision making problem. It can be approached through a number of formalisms familiar to the ML community, including reinforcement learning \autocite{Gottipati2020-hw}, imitation learning \autocite{Mollard2015-qb}, evolutionary design \cite{Peysakhov2003-sg,Devert2006-re}, and Bayesian optimization \cite{Kim2020-cg}. We are unaware of the use of deep generative models for LEGO. However, generative graph models (GGMs), an emerging sub-field of graph representation learning \cite{Hamilton2017-ry}, are particularly well suited to this domain. 



LEGO structures can be represented as a graph, with nodes representing bricks and edges representing connections between bricks. Nodes can hold information, such as brick type, orientation, and colour. Edges can hold information such as how the bricks connect to one another. Graph-based representations of LEGO have been used as a search space for evolutionary design \cite{Peysakhov2003-sg}, and these should be learnable by GGMs such as  Deep Generative Models of Graphs (DGMG) \cite{Li2018-hw} or GraphRNN \autocite{You2018-qf}.\looseness=-1

This paper explores GGMs for designing LEGO structures that are ``human-like'' in their build quality. It makes the following contributions:
\begin{itemize}
    \item We make a case for the use of deep generative models in sequential assembly. We propose LEGO as a platform for experimentation that balances accessibility and realism.  
    \item We propose a model based on  \cite{Li2018-hw} incorporating physical constraints designed for and evaluated on LEGO but general enough for other modular assembly applications.
    \item We perform an extensive evaluation of various metrics proposed in the generative modeling of images, using a novel permutation analysis.
\end{itemize}


\section{Background}
\label{sec:background}

The problem we consider in this paper is the creation of novel "human-like" assemblies expressed as graphs. Here, we briefly review the works most relevant to our problem domain and methodology.

\subsection{LEGO modelling}

LEGO has received much attention within the computer graphics community. One of the most frequently studied LEGO-related problems is finding a constructable and stable layout of bricks for a target object, so-called ``legolization'' \autocite{Testuz2013-tb,Luo2015-qr,Zhou2019-qh}. Legolization techniques \autocite{Kim2014-iu} typically generate an assembly for a given 3D design. In the generative setting, we aim to create \emph{novel} builds which may be conditioned on context, for example a textual description or object class, but not a specific target. This makes evaluation much more challenging because we are not measuring the discrepancy between a target build and the model's output. Instead, we must assess the generative model based on the \emph{quality} and \emph{diversity} of its builds, as well as the \emph{consistency} with any conditioning information.

Another community that has considered LEGO as a problem domain is evolutionary computing. These works are concerned with creative outputs that respect certain structural and aesthetic properties. Relevant to our work, several papers design new representations for LEGO that can be optimized. \textcite{Devert2006-re} introduce a representation of construction plans for LEGO-like structures that has several desirable properties such as re-usability and modularity. \textcite{Peysakhov2003-sg} utilize a graph representation and Genetic Algorithms to create LEGO structures.

In computer vision, \textcite{Jones2019-nt} recover a 3D LEGO model of an assembly from a video of it being assembled. Like this work, we model LEGO designs as a sequential process. In contrast, we work with a more direct representation of structures and do not attempt to learn through vision. Most relevant to our work is \autocite{Kim2020-cg}, which applies Bayesian optimization to the sequential assembly of LEGO structures with generation conditioned on a high-level description or a specific training example. This is extremely similar to the problem we explore.

\subsection{Generative graph models}
Traditional graph generative models are based on random graph theory that formalizes a simple stochastic generation process and have well-understood mathematical properties. For example, Kronecker graphs \cite{leskovec2010kronecker} build up graphs from a small base generator graph by iteratively applying the Kronecker product. The resulting graphs are, by construction, very self-similar due to their simplicity and therefore only capture a few graph statistics such as degree distribution. These graph models remain highly-constrained in terms of the graph structures they can represent. 

Recent graph generative models use neural networks to capture the distribution over random graphs. The quality of graph generative modeling depends on learning the distribution given a collection of relevant training graphs. A number of deep generative models are based on variational autoencoders (VAEs). For instance, the GraphVAE algorithm \cite{simonovsky2018graphvae} uses a VAE to learn a matrix of edge probabilities between every possible pair in a graph. 
An extension of GraphVAE, the Regularized Graph VAE, uses validity constraints to regularize the output distribution of the decoder. Although these models have greater capacity to learn structural information from data than the traditional models, capturing a set of specific global properties such as graph connectivity, and node compatibility is challenging.

A few models generate a graph sequentially by choosing nodes and edges step-by-step. For instance, \textcite{dai2018syntax} use recurrent neural networks (RNNs) to model graph formation, and to capture semantic validity, attribute grammars are applied on the output distribution. 
Deep Generative Models of Graphs (DGMG) \autocite{Li2018-hw} employs a RNN to make a sequence of decisions: whether to add a new node, which node to add, whether to add an edge, and which destination node to connect to the new node. DGMG assumes the probability of a graph is the sum over all possible node permutations. 
GraphRNN \cite{You2018-qf}, uses a recurrent neural network to obtain the distribution over the $i$th row of the lower triangle of the adjacency matrix, conditioned on the previous rows. Nodes and edges are generated with a graph-level RNN which adds a new node and an edge-level RNN which generates a connection between this new node and the nodes of the graph. 

Domain-specific methods such as molecular graphs have recently attracted a lot of attention in drug discovery and material science \autocite{de2018molgan,jin2018junction,li2018multi}.
We believe that mixing domain-specific knowledge and machine learning in physical design will yield similar successes.

\subsection{Evaluation of graph generative models}
In general, evaluating generative models of graphs is challenging due to their complex structural dependencies \cite{theis2015note}. 
Recent relevant graph generative models \cite{You2018-qf, Li2018-hw} evaluate the quality of generation by considering a set of graph statistics: the degree distribution, clustering coefficient and number of occurrence of all orbits with $n$ nodes. To estimate the similarity between generated graphs and the ground truth, the maximum mean discrepancy (MMD) is calculated for each of these statistics using the Wasserstein distance. MMD determines whether two sets of samples from the distribution $p$ and $q$ are derived from the same distribution. However, MMD works on fairly small graphs and for medium-sized graphs such as those we encounter in LEGO, the computation of MMD is very slow due to the Gaussian EMD kernel. 

Another metric that measures the distance between generated and actual samples is the Fr\'echet Inception Distance (FID) \cite{Heusel-fid}. The FID was originally introduced to measure the quality of generated image samples. FID along with other common GAN evaluation metrics use the pretrained image classifier Inception v3 \cite{Szegedy-inceptionv3} to obtain feature representations of images, which enables a more straightforward comparison between generated and reference distributions \cite{Narmm-PRDC, Heusel-fid, Salimans-InceptionScore, kynkaanniemi2019improved, Binkowski-KID}. 
\textcite{liu_auto-regressive_2019} adapt FID to the graph domain by replacing Inception v3 with a Graph Isomorphism Network (GIN) classifier \cite{Xu-GIN}. \autocite{liu_auto-regressive_2019} also introduces GIN classifier accuracy, which is simply the percentage of class conditional samples that can be successfully recognized by the GIN classifier.

\section{Methodology}
Here we present the Deep Generative Model of LEGO Graphs (\alg), a sequential generative model of LEGO structures. We first define a representation that allows us to convert a LEGO structure to a graph. Then, we define the graph generation process for LEGO and show how it is trained. Finally, we adapt a number of generative evaluation metrics to the graph setting.
\subsection{LEGO structure graph representation} \label{sec:graph-rep}
\begin{figure}[t!]
    \centering
    \begin{subfigure}[b]{0.45\textwidth}\centering
        {\includegraphics[width=0.5\textwidth]{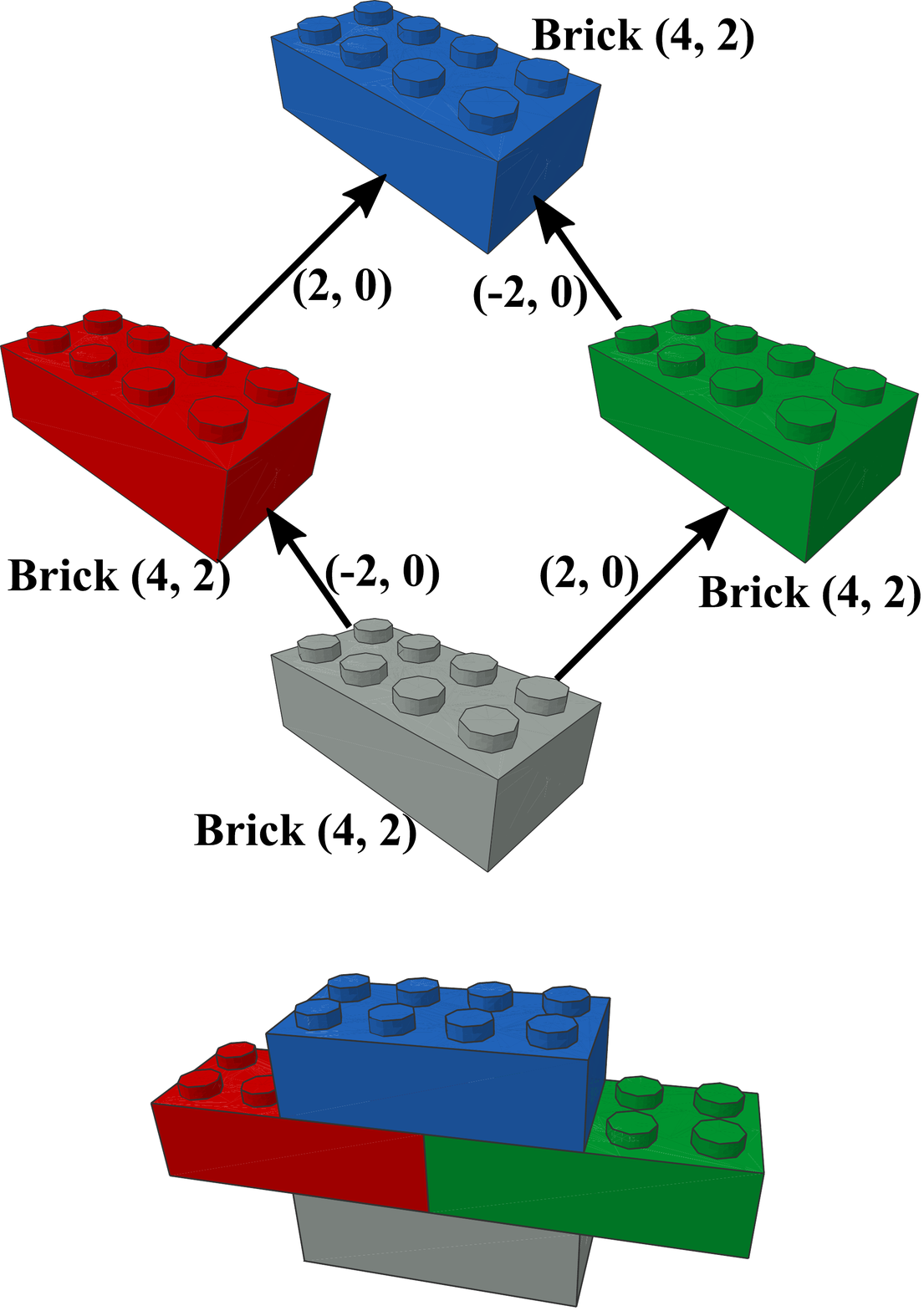}}
        \caption{A valid graph and its corresponding LEGO rendering}
        \label{subfig:valid-graph}
    \end{subfigure}
    \begin{subfigure}[b]{0.5\textwidth}
        \begin{subfigure}{\textwidth}
        \vspace{-4cm}
            \centering
            \includegraphics[width=0.5\textwidth]{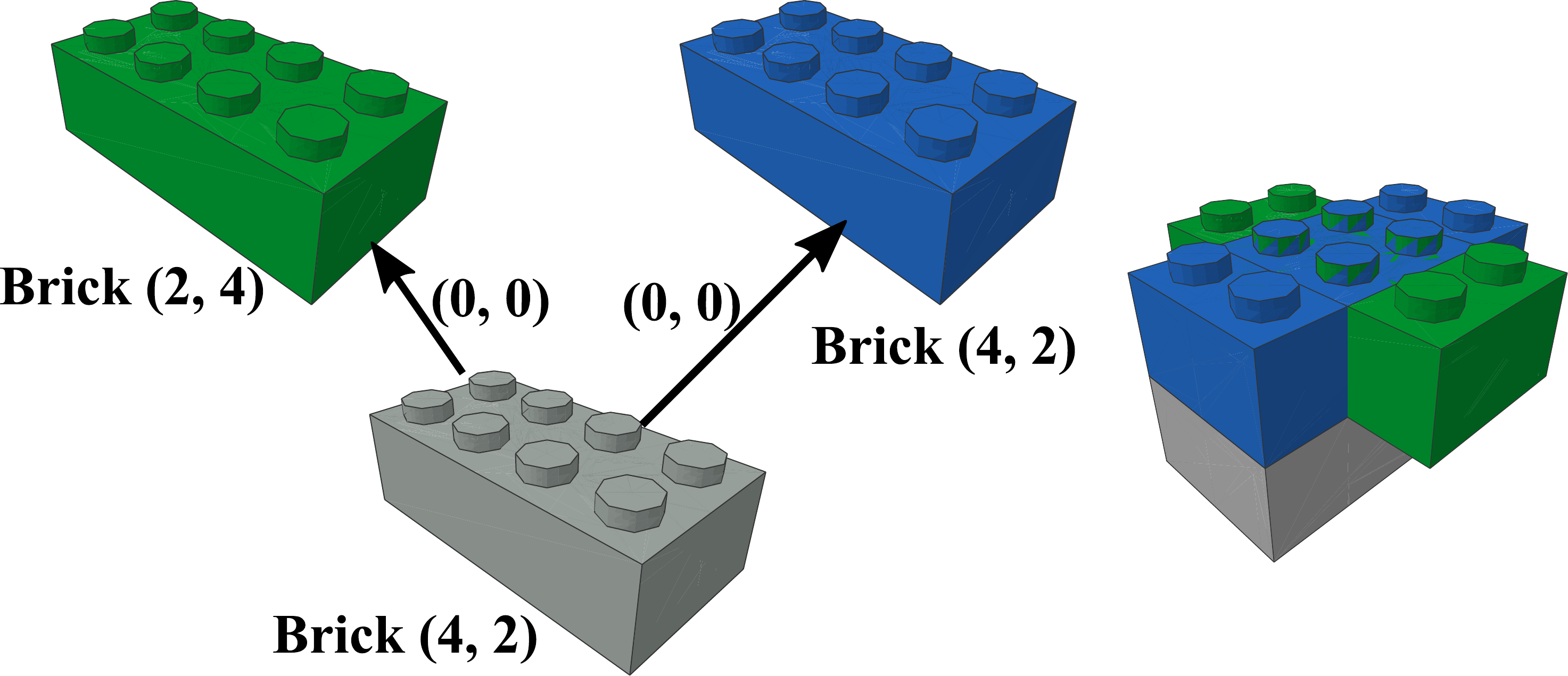}
            \caption{An example of two bricks occupying the same space and its corresponding graph representation}
            \label{subfig:merged}
        \end{subfigure}
        \begin{subfigure}[b]{\textwidth}
            \centering
            \includegraphics[width=0.5\textwidth]{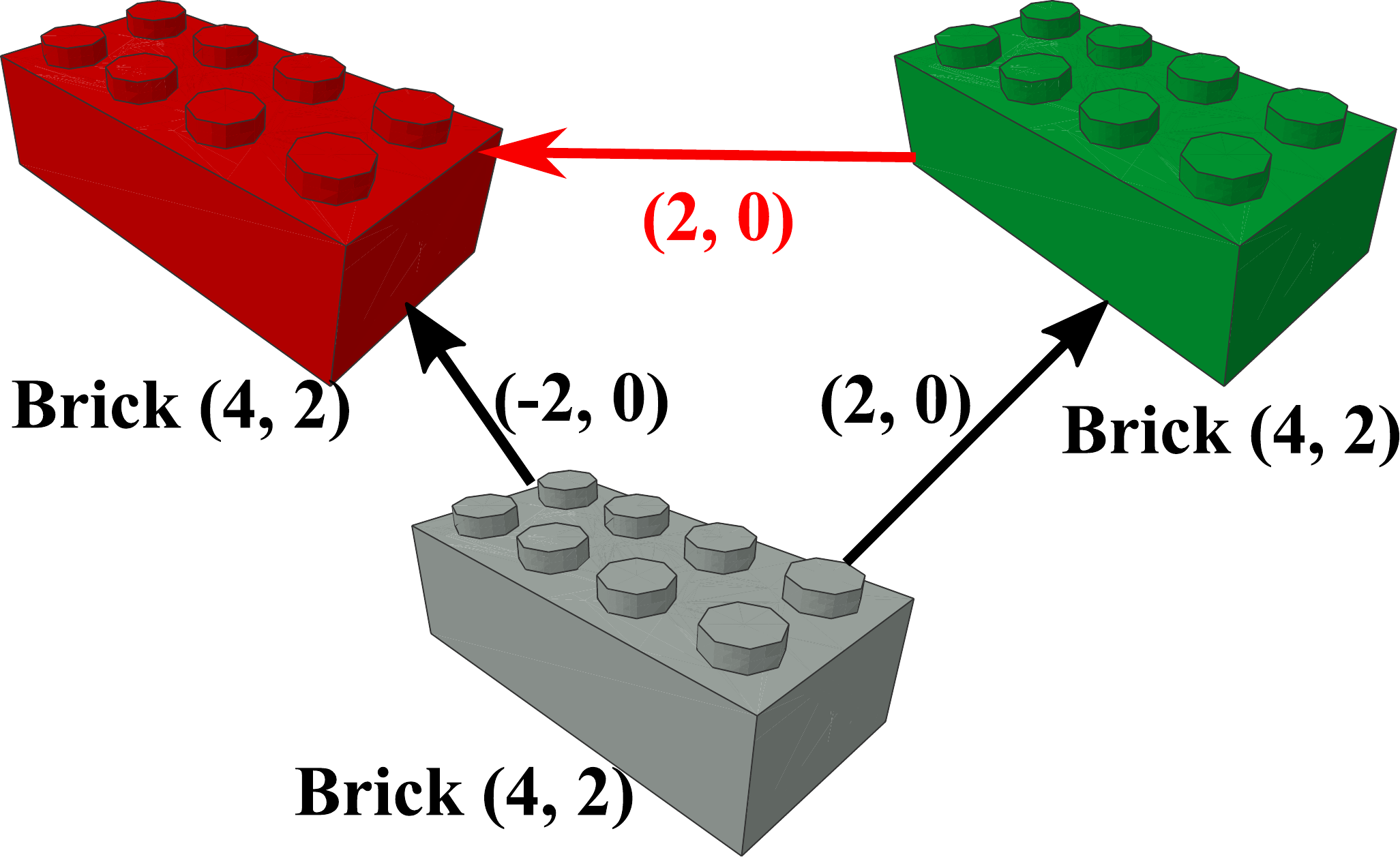}
            \caption{An example of impossible connections}
            \label{subfig:overconstrained}
        \end{subfigure}
    \end{subfigure}
    \caption{Valid and invalid LEGO builds and their graph representations.}
    \label{fig:legograph}
\end{figure}

We utilize a graph representation very similar to the one proposed by Peysakhov et al. \cite{Peysakhov2003-sg}, with small changes to improve compatibility with our generative model. LEGO structures that contain standard symmetrical pieces such as bricks or plates can be represented by this graph representation, however structures with more complex pieces such as wheels or axles cannot. A LEGO structure that meets this criteria may be represented by a directed and labeled graph $G$, where the nodes represent LEGO bricks and the edges represent connections between bricks. In addition, node labels encode the orientation and size of each brick, and edge labels specify how two bricks are connected by encoding a two-element offset ($x$,$y$) between them. A directed edge from node $u$ to node $v$ indicates that brick $u$ provides studs to brick $v$ (i.e. $v$ sits on $u$). 
Although flexible, our graph representation does not ensure that all graphs are physically realizable as LEGO structures. It is possible to form a graph where any two bricks occupy the same physical space, or to overconstrain a brick to be in multiple locations at once. Invalid graphs are not physically realizable because of one or more of the discussed issues. An example of these issues along with a valid LEGO graph is shown in Figure \ref{fig:legograph}.

An idiosyncrasy with our graph-based representation of LEGO structures is that some edges are redundant. An edge connecting $u$ to $v$ may imply that $u$ is also connected to another node $k$. We refer to this as an ``implied edge'', and an example can be found in Figure \ref{subfig:valid-graph}; if any edge is removed the underlying LEGO structure remains unchanged, and the removed edge becomes implied. Generated graphs are considered to be valid regardless of any missing implied edges.

\subsection{Sequential graph generation}\label{sec:graphgen}
We expand upon the DGMG graph generation process \cite{Li2018-hw} to create a model that is capable of making decisions regarding typed and directed edges. Wherever possible we modelled our extensions after the sequence of actions a human might follow when creating a LEGO structure.
DGMG uses a sequential process to generate nodes one at a time and connect them to the existing partial graph. At each iteration, it determines whether a new node of a particular type should be added, or the generation process should terminate. If a new node is added, DGMG then chooses whether to add an edge to this node or not. A node in the existing graph is then chosen as a destination for a newly added edge. This edge generation process is repeated until the decision is made to stop connecting new edges, in which case the process continues from the node generation step. The entire process is repeated until the node generation step makes the decision to terminate. DGMG uses graph-structured neural networks (graph nets) to complement structure building decisions by using message passing and graph propagation to yield node and edge representations \autocite{Hamilton2017-ry}. 

\paragraph{Graph propagation} For a graph $G = (V, E)$, we associate a node embedding vector $\mathbf{h}_v \in \mathbb{R}^H$ for all $v \in V$, and an edge embedding vector $\mathbf{s}_{u,v} \in \mathbb{R}^S$ for all $e \in E$. The set of all node embeddings in G is denoted by $\mathbf{h}_V = \{h_1, h_2, \ldots, h_{\lvert V \rvert}\}$. The embeddings are initialized using corresponding node and edge types, and are used in the graph propagation process to aggregate information across the graph. As in \autocite{Li2018-hw}, the function $f_e$ that computes the message vector $a_v$ from $u$ to $v$ is a fully-connected network, and the node-update function $f_n$ is a gated recurrent unit (GRU) cell:

\begin{minipage}{\textwidth}
    \noindent\begin{minipage}{.5\linewidth}
        \begin{equation}
            \label{eq:message_vector}
            \mathbf{a}_v = \sum\limits_{u:(u,v)\in E} f_e(\mathbf{h}_u, \mathbf{h}_v, \mathbf{s}_{u,v}) \quad\forall v\in V,
        \end{equation}
    \end{minipage}
    \begin{minipage}{.5\linewidth}
        \begin{equation}
            \label{eq:node_update}
            \mathbf{h'}_v = f_n(\mathbf{a}_v, \mathbf{h}_v) \quad\forall v \in V .\
        \end{equation}
    \end{minipage}
\end{minipage}

\textcite{Li2018-hw} suggest using a different set of parameters for $f_e$ and $f_n$ for each round of graph propagation to increase model capacity, and we use this setting. The function prop$^{(T)}(\mathbf{h}_V, G)$ denotes $T$ rounds of graph propagation, and returns a set of updated node embeddings $\mathbf{h}_V^{(T)}$. This is equivalent to repeating Eq.~\ref{eq:message_vector} and Eq.~\ref{eq:node_update} $T$ times. We set $T$ to 2 throughout all experiments:
\begin{equation}
    \label{eq:graphprop}
    \mathbf{h}_V^{(T)} = \textrm{prop}^{(T)}(\mathbf{h}_V, G) .\
\end{equation}
The new node vectors $\mathbf{h}_V^{(T)}$ are carried through the decision modules below, making them recurrent across these decisions. The node vectors are also recurrent across the graph propagation steps.

To obtain graph embeddings, we first map the node representations to a higher dimension using a fully connected network $f_m$: $\mathbf{h}_v^G = f_m(\mathbf{h}_v)$. We then apply a gated sum over all nodes to obtain a single vector $\mathbf{h}_G$. The function $g_m$ is a fully collected network which maps each node embedding to a single value, and determines $\mathbf{g}_v^G$, the importance of each node, for use in the gated sum:

\begin{minipage}{\textwidth}
    \noindent\begin{minipage}{.5\linewidth}
        \begin{equation}
            \label{eq:graphembed}
            \mathbf{h}_G = \sum\limits_{v \in V}\mathbf{g}_v^G\odot\mathbf{h}_v^G,
        \end{equation}
    \end{minipage}
    \begin{minipage}{.5\linewidth}
        \begin{equation}
            \label{eq:node-importance}
            \mathbf{g}_v^G = \sigma(g_m(\mathbf{h}_v)) .\
        \end{equation}
    \end{minipage}
\end{minipage}

\paragraph{Add node module} In this module, we produce the probability of adding a node of each type and the probability of terminating the process using an existing graph $G$ and its corresponding node embeddings $\mathbf{h}_V$. We first use Eq.~\ref{eq:graphprop} to run $T$ rounds of graph propagation to obtain updated node vectors $\mathbf{h}_V^{(T)}$, which are then used to create a graph representation vector as in Eq.~\ref{eq:graphembed}. The graph embedding vector is then passed through a standard MLP $f_{an}$ with softmax activation to obtain the probability associated with each possible action:
\begin{equation}
    \label{eq:addnode}
    f_{addnode}(G) = \textrm{softmax}(f_{an}(\mathbf{h}_G)) .\
\end{equation}
\paragraph{Add edge module} In the add edge module, we take an existing graph $G$, a newly added node $v$, and compute probabilities for three possible outcomes: not adding an edge to $v$, adding an incoming edge to $v$, or adding a outgoing edge from $v$. These values are determined by passing the graph representation vector $\mathbf{h}_G$ and the new node embedding $\mathbf{h}_v$ through another MLP $f_{ae}$ with softmax out:\looseness=-1
\begin{equation}
    \label{eq:addedge}
    f_{addedge}(G, v) = \textrm{softmax}(f_{ae}(\mathbf{h}_G,\mathbf{h}_v)).
\end{equation}
\paragraph{Choose destination module} This module computes a score $x_u$ for every $u \in V\setminus \{v\}$ using an MLP $f_s$, and normalizes the vector $\mathbf{x}$ through a softmax to obtain the probability of connecting the new node $v$ to $u$ with direction determined by the add edge module:
\begin{minipage}{\textwidth}
    \noindent\begin{minipage}{.5\linewidth}
        \begin{equation}
            \label{eq:choosedestscores}
            x_u = f_s(\mathbf{h}_u^{(T)},\mathbf{h}_v), \quad \forall u \in V \setminus v,
        \end{equation}
    \end{minipage}
    \begin{minipage}{.5\linewidth}
        \begin{equation}
            \label{eq:choosedest}
            f_{dest}(G, v)=\textrm{softmax}(\mathbf{x}).\
        \end{equation}
    \end{minipage}
\end{minipage}

This module may handle typed or directed edges by making $x_u$ a vector of scores the same size of edge types \cite{Li2018-hw}. However, we separate these decisions to avoid combinatorial explosion in the number of outputs. The arguments to $f_s$ in Eq.~\ref{eq:choosedestscores} are rearranged such that the source node and destination node for a newly added edge are always given to the MLP in the same order.
\paragraph{Choose edge type module} This module determines the edge type of the newly added edge between nodes $u$ and $v$ by choosing the ($x$,$y$) offset between bricks $u$ and $v$. We treat this decision as two independent events, and use two separate MLPs $f_{ex}$ and $f_{ey}$ to determine the x and y offset, respectively: 
\begin{minipage}{\textwidth}
    \noindent\begin{minipage}{.5\linewidth}
        \begin{equation}
            \label{eq:xshift}
            x = 
              \textrm{softmax}(f_{ex}(\mathbf{h}_u^{(T)}, \mathbf{h}_v)),
        \end{equation}
    \end{minipage}
    \begin{minipage}{.5\linewidth}
        \begin{equation}
            \label{eq:yshift}
            y = 
              \textrm{softmax}(f_{ey}(\mathbf{h}_u^{(T)}, \mathbf{h}_v)).\
        \end{equation}
    \end{minipage}
\end{minipage}

We experiment with two ways to pose this problem: i) as an ordinal regression problem where each output has a clear rank \cite{Cheng-ordinal-regression}, or ii) by treating the offsets as categories, as in classification. The former is described in Appendix \ref{sec:ordinal-regression}, while the latter is shown in Eq.~\ref{eq:xshift} and Eq.~\ref{eq:yshift}. In addition, this module introduces a new type of invalid graph; although two bricks may be connected in the graph, it is possible that we generate an offset such that it is physically impossible for them to connect.

 In summary, to generate a graph for the LEGO graph representation: (1) choose whether to add a LEGO brick with a given size and orientation, or terminate the assembly process, (2) choose whether to attach this new brick into the LEGO structure, and if so whether it should connect on top of or underneath of a pre-existing brick, (3) if the newly added brick is to be connected, decide which brick it should connect to, otherwise go back to step (1), and (4) choose the relative offset of these bricks from one another, and restart from step (2). We perform class-conditioned generation by adding a ``one-hot'' class-conditioning vector $\mathbf{c}$ to the input of each structure building MLP described above.
 

\subsection{Training and evaluation}
Given a set of training graphs, we train our model to maximize the joint log-likelihood $\mathbb{E}_{P_{data}(G)} \log p(G)$ using categorical cross-entropy. For each LEGO structure, a graph generating sequence is created knowing that the ordering is analogous to an assembly that a human would make.  The likelihood for each individual step is computed using the output modules described in \S~\ref{sec:graphgen}.\looseness=-1

To evaluate the generated structures, we adapt several popular and thoroughly tested GAN evaluation metrics to the generative graph domain. Most are designed for images and make use of a pretrained Inception v3 network as a fixed feature extractor. As in \autocite{liu_auto-regressive_2019}, we replace Inception v3 with a Graph Isomorphism Network (GIN) \autocite{Xu-GIN} to obtain feature representations of graphs\footnote{When we generalize a method beyond Inception v3, we will simplify its acronym. For example FID \textrightarrow FD. To avoid introducing another acronym, in \S~\ref{sec:experiments} we use the simplified acronym because our embeddings are unambiguously GIN.}.

\section{Experiments}
\label{sec:experiments}
We execute two main experiments. The first demonstrates the effectiveness and value of the \S~\ref{sec:new-metrics} metrics for generative LEGO assembly. The second evaluates our model quantitatively with these metrics, and qualitatively by visualizing generated LEGO structures. 
In both experiments we use the LEGO dataset from \textcite{Kim2020-cg}. This dataset consists of 12 classes and a total of 360 LEGO structures built using 2\texttimes4 LEGO bricks, with each structure created by one of twelve human subjects. More information regarding the dataset is provided in Appendix \ref{sec:dataset}.

\subsection{Permutation analysis}
This experiment demonstrates the value of our proposed metrics in the evaluation of generative graph models. We begin by duplicating the LEGO dataset and designating the original as an unchanging reference. Throughout the experiment, we apply several stacking permutations to each graph in the copy. At each iteration every graph in this copy is randomly permuted, and the changes are carried over to the next iteration. These permutations cause the distribution of the permuted dataset to slowly drift from the reference distribution. We expect the drift to be detected by all evaluation metrics. 

The permutations applied are simple: with equal probability, a randomly selected node and all associated edges are deleted from the graph, or a node with a randomly selected brick type is added to the graph. To ensure some similarity to the reference dataset at a very high level, node deletions that result in a disjoint graph are prohibited. To complete the addition of a node, we form a connection to a random pre-existing node, and assign a random edge direction and type under the constraint that the resultant graph must represent a valid LEGO structure. Once a valid edge is created, all implied edges are determined and subsequently added to the graph. The constraints ensure that the permuted dataset will resemble randomly assembled LEGO structures in both the LEGO and graphical representations, and allows for the results to be evaluated both qualitatively and quantitatively. We perform 500 accumulating permutations to each graph in the copy. 

\begin{figure}
    \centering
    \begin{subfigure}[b]{\textwidth}
        \hfill
        \includegraphics[width=0.95\textwidth]{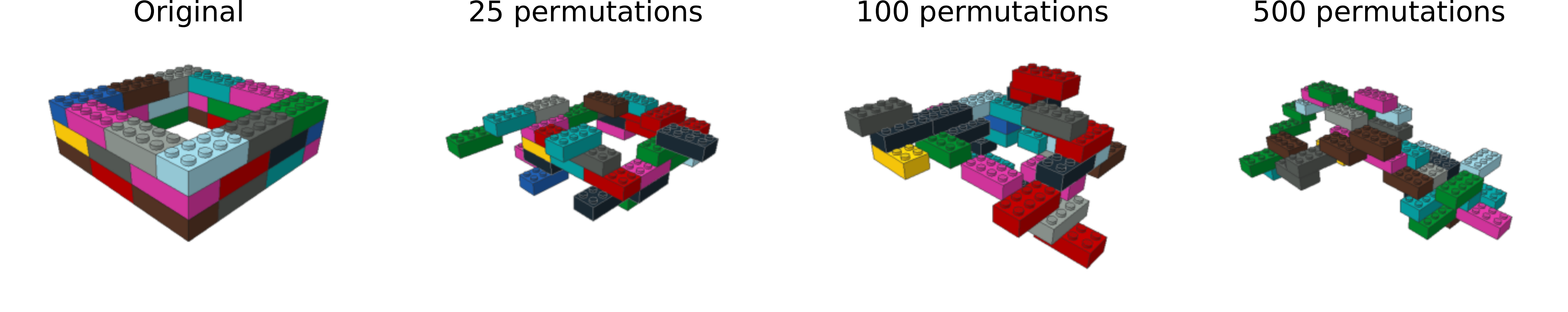}
    \end{subfigure}
    \begin{subfigure}[b]{\textwidth}
        \centering
        \vspace{-0.35cm}
        \includegraphics[width=\textwidth]{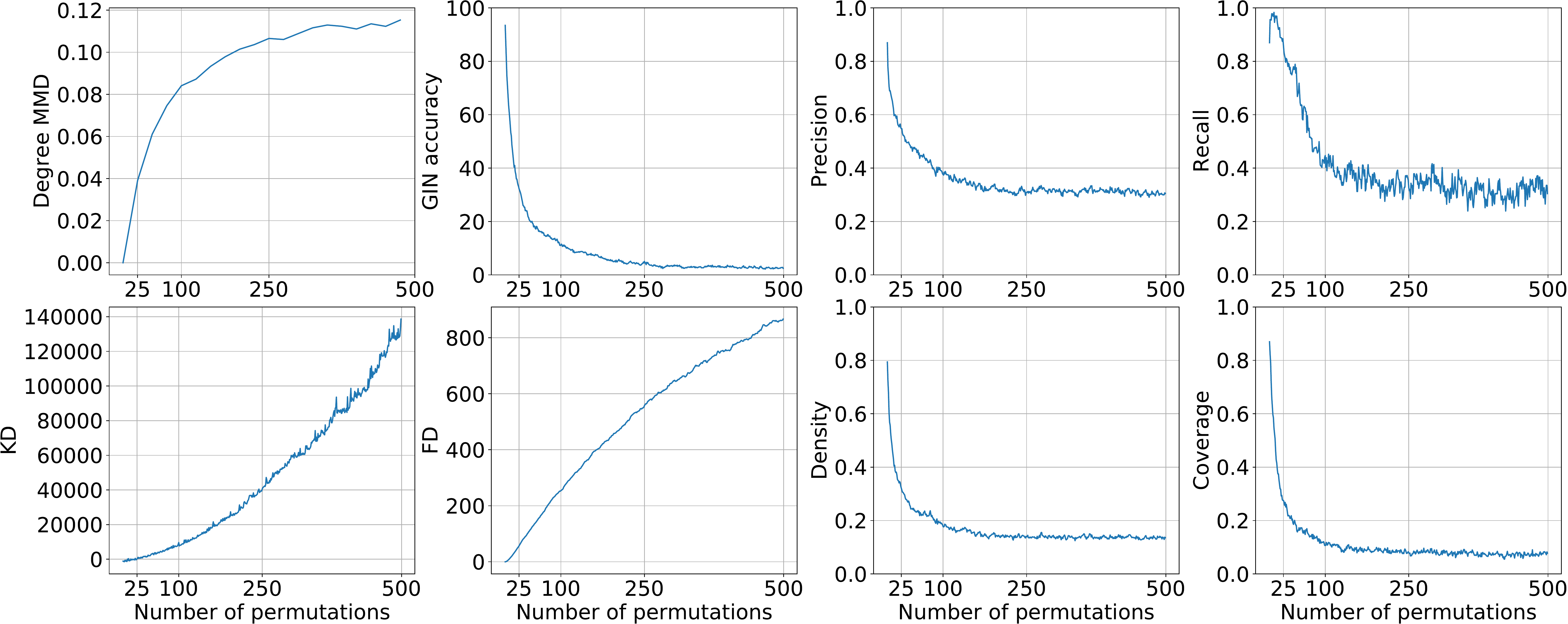}
    \end{subfigure}
    \caption{An example of a LEGO structure as it undergoes several permutations (top), and plots of various metrics as stacking permutations are applied (bottom). All GIN-based metrics are calculated at every iteration while degree MMD is calculated every 25 iterations due to its computational cost.}
    \label{fig:graph-permutation-plot}
\end{figure}

We can see from Figure \ref{fig:graph-permutation-plot} that all metrics included in the experiment capture the distribution shift relatively well. It is expected that each LEGO structure will reach a point where it appears that it was randomly assembled, and the metrics should then begin to asymptote. This is the case for all metrics with the exception of FD and KD; FD appears to asymptote, while KD is almost exponential, indicating a flaw may be present in the metric. GIN accuracy, density, and coverage decline extremely quickly and appear to be sensitive to any sort of distribution shift. With the exception of KD and recall, all metrics are fairly smooth with little noise. We include degree MMD from \textcite{You2018-qf} in the experiment, and it is also capable of capturing the distribution shift.

\newcommand{\BOclass}{{BO-CC}\xspace}
\newcommand{\BOinstance}{{BO-SI}\xspace}
\subsection{Generation}
Next we train our class-conditioned generative model on the same dataset considered in the permutation analysis. In addition, we employ the Bayesian optimization sequential LEGO assembly method from \autocite{Kim2020-cg} as a baseline for our generative experiments. That method employs a voxel representation of LEGO structures, and restricts the decision space to prevent the creation of invalid assemblies. Given a partially assembled LEGO structure, the position of the next LEGO brick is posed as a score maximization problem, and Bayesian optimization is utilized to efficiently select positions to evaluate. The authors describe two different evaluation functions  --- one that uses an entire class of training examples, and one that only uses a single example. The latter is an easier problem and is how the model was designed to be used, and thus should yield better results, while the former is more comparable to our problem formulation. We employ both methods as baselines in our experiments.

\begin{figure}
    \centering
    \includegraphics[width=0.75\linewidth]{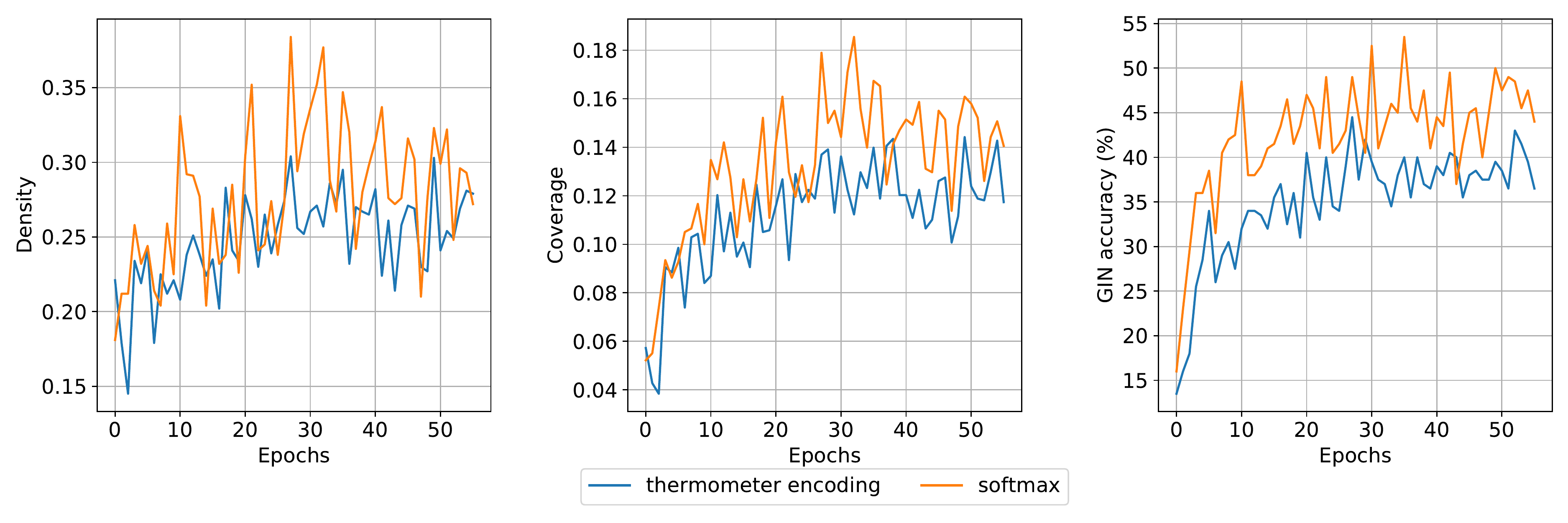}
    \caption{A comparison of density, coverage, and GIN accuracy metrics for two variants of \alg, which employ different methods for determining edge types.}
    \label{fig:ordinal-classification}
\end{figure}

First, we compare two methods for configuring the edge type module: using a thermometer encoding as motivated by ordinal regression \cite{Cheng-ordinal-regression} where the integer nature of the offsets is explicit, and as a softmax where each offset is treated as a category and thus maximally different. Surprisingly, the latter seems to provide a small increase to stability during training and provides slightly better results across the majority of metrics as shown in Figure \ref{fig:ordinal-classification}.

We compare our model's performance to the baselines using the GIN-based metrics proposed in Section \ref{sec:new-metrics}. We fix the reference dataset to be the entirety of the dataset, and generate 200 samples with each method to prevent any discrepancy caused by differing sizes. We use the harmonic mean of density and coverage to identify the model with the highest performance through all epochs. We have found FD and KD to be relatively noisy with only 200 generated samples, and similar to the unified metrics, the harmonic mean of D\&C provides a balance between sample quality and diversity. 

We report all proposed metrics in Table \ref{tab:generative-results} for two different versions of this model: the unrestricted variant (\alg), and one where we prevent any decisions that would result in an invalid structure (\alg-Re). Both versions of \alg perform significantly better than the baselines across most GIN-based metrics. As noted above, our models and \BOclass are conditioned on class while \BOinstance is conditioned on individual training examples, so the most appropriate comparison is among \alg, \alg-Re and \BOclass. An interesting result is that \alg outperforms \alg-Re. We expected that limiting the generated graphs to physically realizable LEGO structures would improve performance due to increased resemblance to the training set. This may be a result of the difference in average size of the generated graphs; \alg averages 52 nodes, \alg-Re averages 78 nodes, and the training set averages 56 nodes. 
It is unclear whether more training, or reducing the maximum size of generated graphs will improve the performance of \alg-Re relative to \alg. 

\begin{figure}
\newcommand{\size}{0.23\textwidth}
    \centering
    \begin{subfigure}[b]{\linewidth}
        \begin{subfigure}[b]{\size} \vspace{1cm}
            \centering
            \includegraphics[width=\linewidth]{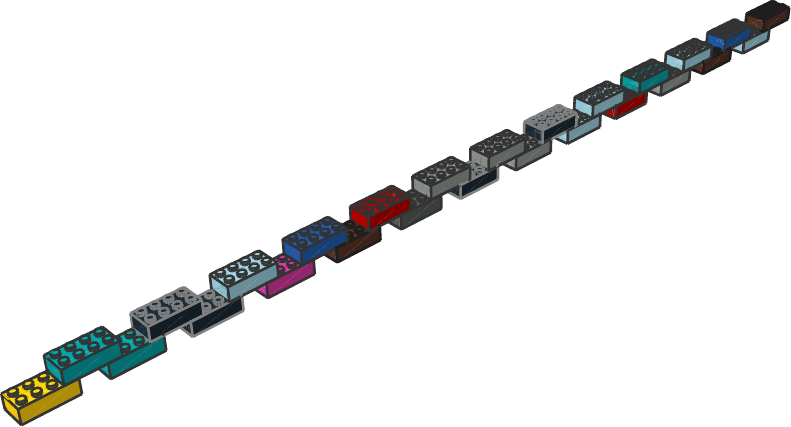}
            \caption{line}
            \label{subfig:line}
        \end{subfigure}
        \begin{subfigure}[b]{\size}
            \centering
            \includegraphics[width=\textwidth]{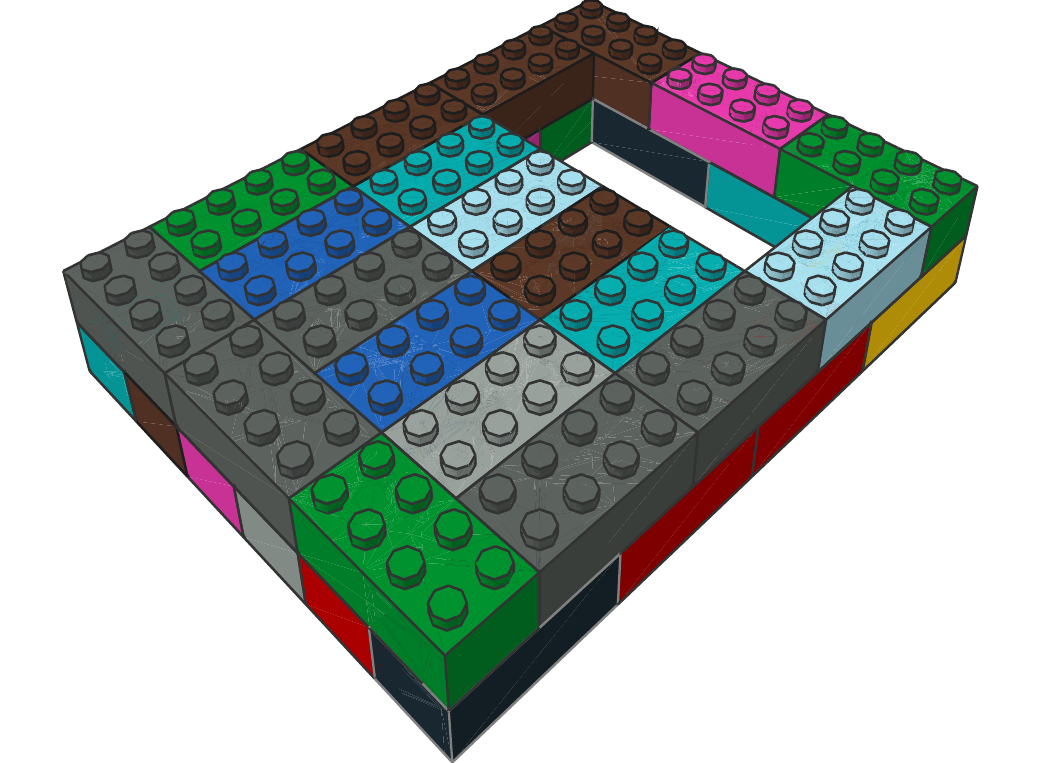}
            \caption{cuboid}
            \label{subfig:cuboid}
        \end{subfigure}
        \begin{subfigure}[b]{\size}
            \centering
            \includegraphics[width=\linewidth]{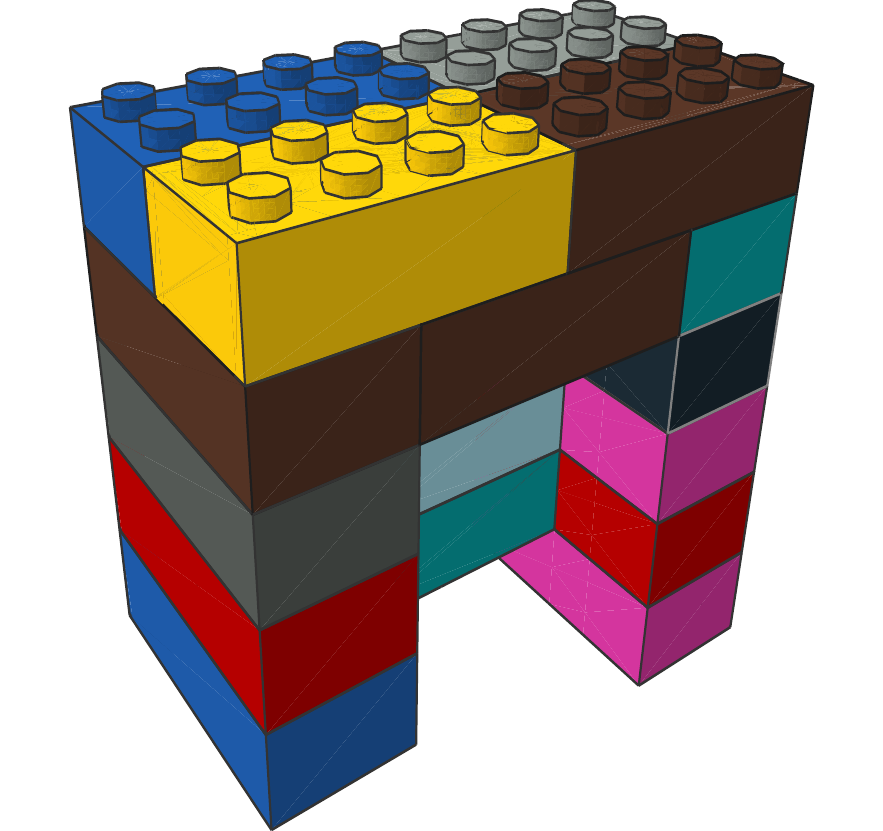}
            \caption{table}
        \end{subfigure}
        \begin{subfigure}[b]{\size}
            \centering
            \includegraphics[width=\linewidth]{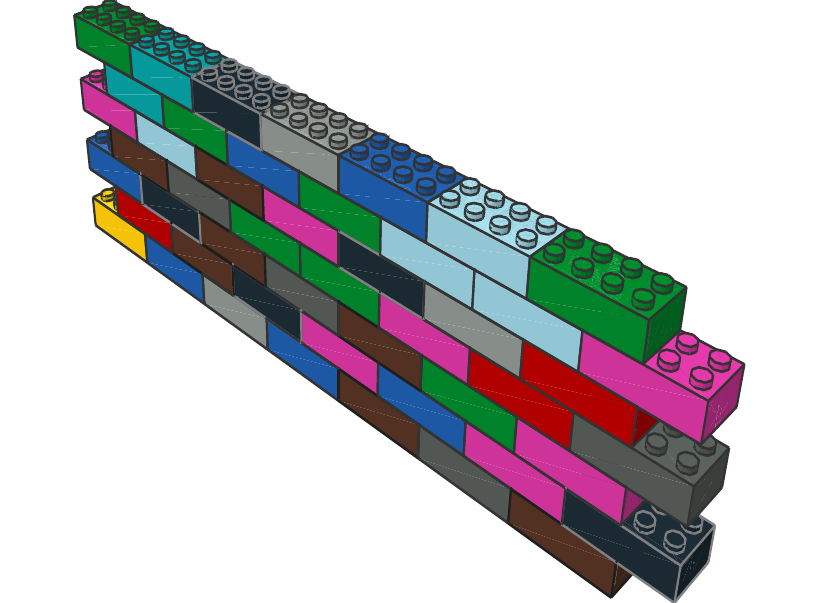}
            \caption{wall}
        \end{subfigure}
    \end{subfigure}
    
    \begin{subfigure}[t]{\linewidth}
        \begin{subfigure}[t]{\size}
            \centering
            \includegraphics[width=\linewidth]{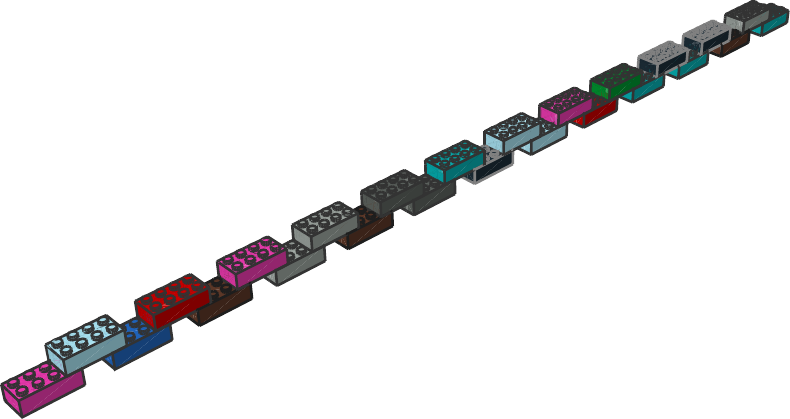}
        \end{subfigure}
        \begin{subfigure}[t]{\size}
            \centering
            \includegraphics[width=\linewidth]{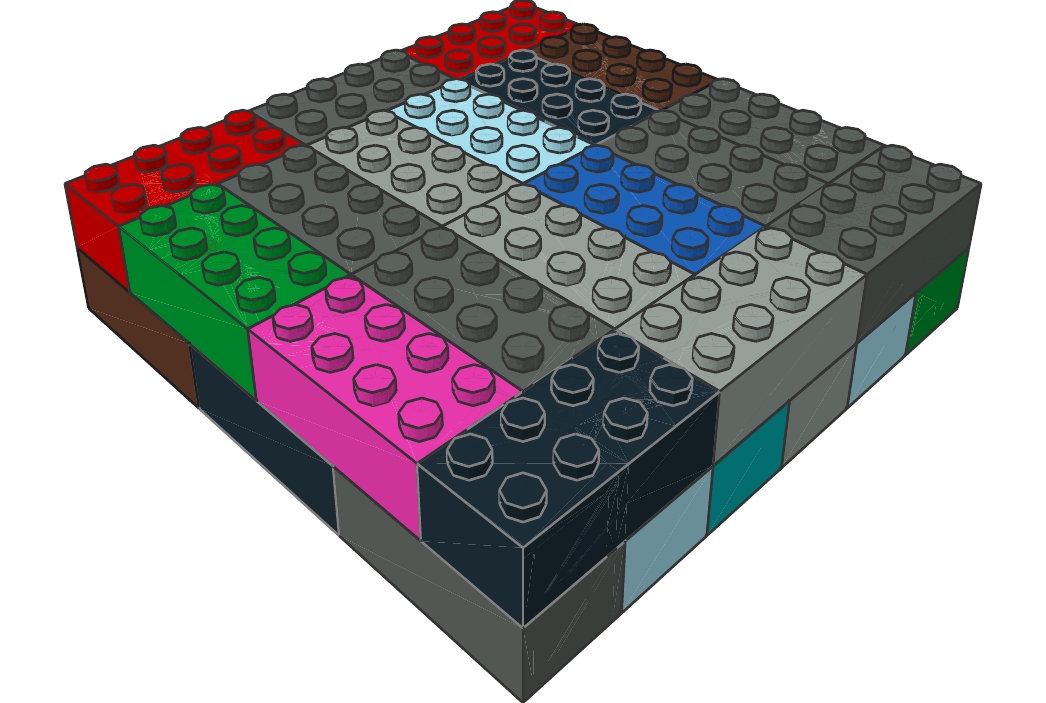}
        \end{subfigure}
        \begin{subfigure}[t]{\size}
            \centering
            \includegraphics[width=\textwidth]{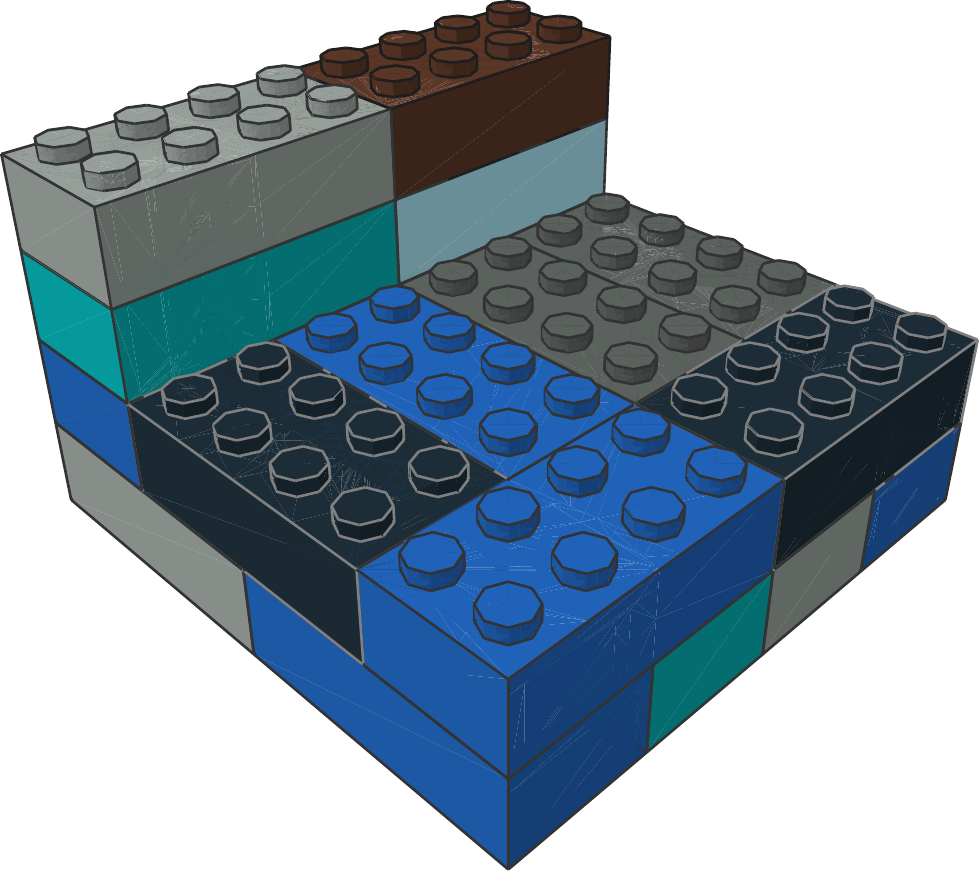}
        \end{subfigure}
        \begin{subfigure}[t]{\size}
            \centering
            \includegraphics[width=\textwidth]{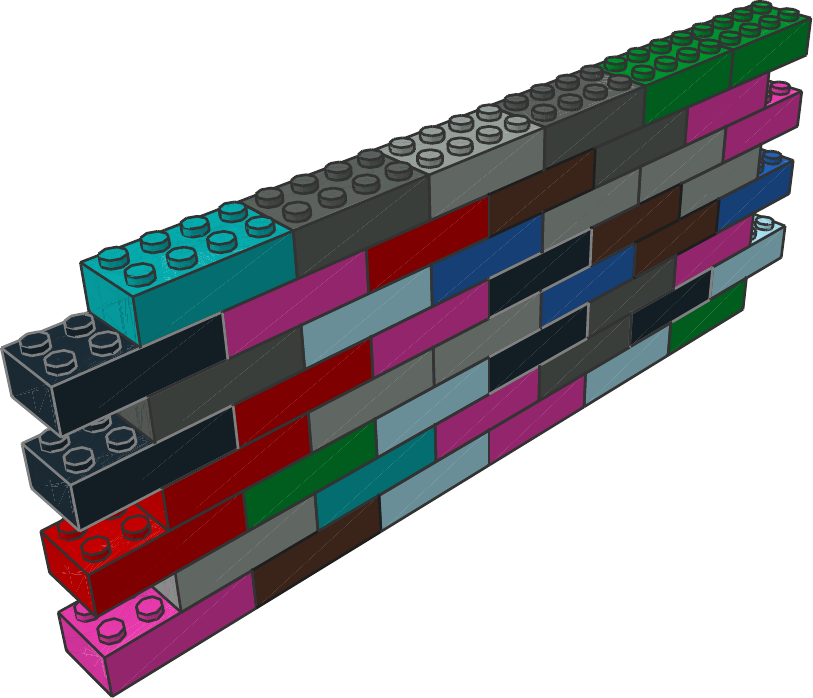}
        \end{subfigure}
    \end{subfigure} 
    \caption{High quality LEGO structures generated by \alg and the class each generation was conditioned on (top), and the corresponding nearest neighbour by GIN graph embedding (bottom). The colour of each brick is randomly selected for aesthetics and not included anywhere in the graph generation process.}
    \label{fig:generated-builds}
\end{figure}

We include visualizations of the LEGO structures generated by \alg, along with each sample's nearest neighbour in the training set to show that the model is not simply generating copies of training examples in Figure \ref{fig:generated-builds}. The generated structures are novel yet share common features with the corresponding class, indicating that \alg is capable of understanding and replicating patterns found in the dataset. For example, the generated structures in Figure \ref{subfig:line} and Figure \ref{subfig:wall} extend patterns to create larger structures than what are found in the dataset. Also, the structure in Figure \ref{subfig:cuboid} shows that the model has learned that this particular class should be perfectly flat without any overhanging bricks. We estimate that these lie in about the top 5\% of the generated structures based on our assessment of visual attractiveness.

\renewcommand{\tabcolsep}{3pt}
\begin{table}
  \caption{Performance evaluation of \alg vs.~the Bayesian optimization baselines on the proposed GIN-based metrics. \BOclass is conditioned on an entire class similar to our work while \BOinstance is conditioned on a single instance as in \autocite{Kim2020-cg}. The \textuparrow/\textdownarrow symbols indicate that higher/lower is better, respectively. GIN Acc = GIN classifier accuracy, P = Precision, R = Recall, D = Density, C = Coverage.\looseness=-1}
  \label{tab:generative-results}
  \centering

  \begin{tabular}{lccccccccc}
    \toprule
    & FD \textdownarrow & GIN Acc \textuparrow & P \textuparrow & R \textuparrow & D \textuparrow & C \textuparrow & KD \textdownarrow & \% novel \textuparrow & \% valid \textuparrow \\
    \midrule
    
    \BOinstance
    \autocite{Kim2020-cg}  &  345  &  24.5 &  0.47  &  0.70 & 0.23 & 0.078  &  5297 & 93.5 &  \textbf{100}\\
    
    \BOclass \autocite{Kim2020-cg}  &  407  &  11.9 &  0.42  &  0.59 & 0.26 & 0.051  &  8901 & \textbf{100} &  \textbf{100} \\
    \alg &  \textbf{150}  & \textbf{60.5} &  \textbf{0.62}  & \textbf{0.92} &  \textbf{0.48}  &  \textbf{0.23}  & \textbf{2054} & 89.5  &  25\\
    
    \alg-Re & 334 & 50.5 & 0.59 & 0.88 & 0.44 & 0.21 & 2.4e4 & 87 & \textbf{100}\\
    \bottomrule
  \end{tabular}
\end{table}

\section{Conclusion} 
We demonstrated that graph generative models can be readily used in sequential assembly of physical structures with visually satisfying results. We also showed the value of adapting several common evaluation metrics popularized by GANs to the generative graph domain. For future real-world physical design, it is vital to test constructability as well as the stability of generated structures to ensure safety. We intend to pursue a larger-scale dataset with a wider variety of brick types and LEGO structures.


\newpage
\printbibliography

\appendix
\newpage
\section*{Appendices}

\section{Evaluation metrics}
\label{sec:new-metrics}

In this section, we describe the specific metrics that we adapt to the graph setting and apply in \S~\ref{sec:experiments}.

\paragraph{GIN Accuracy} A simple measure of generation quality is to determine what percentage of class conditional samples can be successfully recognized by a pretrained classifier. If the generated samples share similar properties with the target class, then the labels predicted by the classifier should match the conditioning labels. The classification accuracy metric was introduced in the GAN literature as GAN-test~\cite{shmelkov2018good}, and has been adapted for evaluating graph generations through the use of a pretrained GIN classifier in \autocite{liu_auto-regressive_2019}. GIN accuracy does not measure sample diversity. As such, a model that only produces a single realistic sample may still achieve a good score.

\paragraph{Fr\'echet Distance (FD)} Fr\'echet Distance may be used to measure the distance between two probability distributions, and is commonly used for evaluating the quality of generative models. Generated samples and reference samples are embedded into some task relevant feature space (e.g.~Inception v3 for images), and a multivariate Gaussian is fit to each set of features. The Fr\'echet Distance between the two distributions can then be calculated as 
$D = ||\boldsymbol{\mu}-\boldsymbol{\hat{\mu}}||^{2}_{2} + \Tr\left(\mathbf{\Sigma} + \mathbf{\hat{\Sigma}} - 2(\mathbf{\Sigma}\mathbf{\hat{\Sigma}})^{1/2}\right)$, where $\mu$ and $\Sigma$ are the mean and covariance of the reference distribution, and $\hat{\mu}$ and $\hat{\Sigma}$ are the mean and covariance of the generated distribution. FD jointly considers both generation realism and diversity, and as such is a good overall measure of model performance. 

\paragraph{Kernel Distance (KD)} Kernel Inception Distance~\cite{Binkowski-KID} was introduced as an alternative to FID that brought with it several advantages: KID does not assume a parametric form of the distribution of the embedding space, it compares skewness in addition to the mean and variance that FID measures, and it is an unbiased estimator. KID can be measured by computing the squared maximum mean discrepancy (MMD) between distributions after a polynomial kernel has been applied.

\paragraph{Precision and Recall (P\&R)}  Precision and Recall~\cite{kynkaanniemi2019improved} were introduced to address FID's coupling of diversity and quality by providing measures that evaluate each property separately. Precision measures generation realism, while Recall measures sample diversity. To measure P\&R with graphs, all real and generated samples are first embedded into a GIN feature space. Manifolds are then constructed by extending a radius from each embedded sample in a set to its $K^{th}$ nearest neighbour to form a hypersphere, with the union of all hyperspheres representing a manifold. Two manifolds are produced: one for real graphs, and one for generated graphs. Precision is defined as the percentage of generated graphs that fall within the manifold of real graphs, while Recall is defined as the percentage of real graphs which fall within the manifold of generated graphs. Though useful, Precision and Recall are susceptible to outliers \autocite{Narmm-PRDC}.

\paragraph{Density and Coverage (D\&C)} Density and Coverage have recently been introduced as robust alternatives for Precision and Recall, respectively ~\cite{Narmm-PRDC}. Density is calculated as the average number of real examples within whose manifold radius each generated sample falls into. Coverage is described as the percentage of real examples which have a generated sample fall within their manifold radius.

\section{Thermometer encoding} \label{sec:ordinal-regression}
We experiment with a thermometer encoding motivated by ordinal regression \cite{Cheng-ordinal-regression} in the edge type module, which accounts for the explicit ordering of each integer offset. In a typical classification problem the goal is to predict the probability that a data point $x$ belongs to its class $k$, and the target vector is $t = (0, 0, \ldots, 1, \ldots, 0, 0)$, where $t_k$ is one and all other elements are zero. This is easily done through the use of a softmax activation function. In the thermometer encoding, if a data point $x$ belongs to class $k$, then it must also be classified into categories $(1, 2, \ldots, k - 1)$. Thus, the target vector becomes $t = (1, 1, \ldots, 1, 0, \ldots, 0)$ where $t_i$ is one if $i \leq k$, and zero if $i > k$. We obtain a prediction $O$ for this vector by independently applying the sigmoid activation function to each output as described by \textcite{Cheng-ordinal-regression}. During generation, we begin with $o_0$ and sample from consecutive elements of $O$ until a zero is obtained.

\section{The LEGO dataset} \label{sec:dataset}
\begin{figure}
    \centering
    \begin{subfigure}[b]{0.3\textwidth}
        \includegraphics[width=\textwidth]{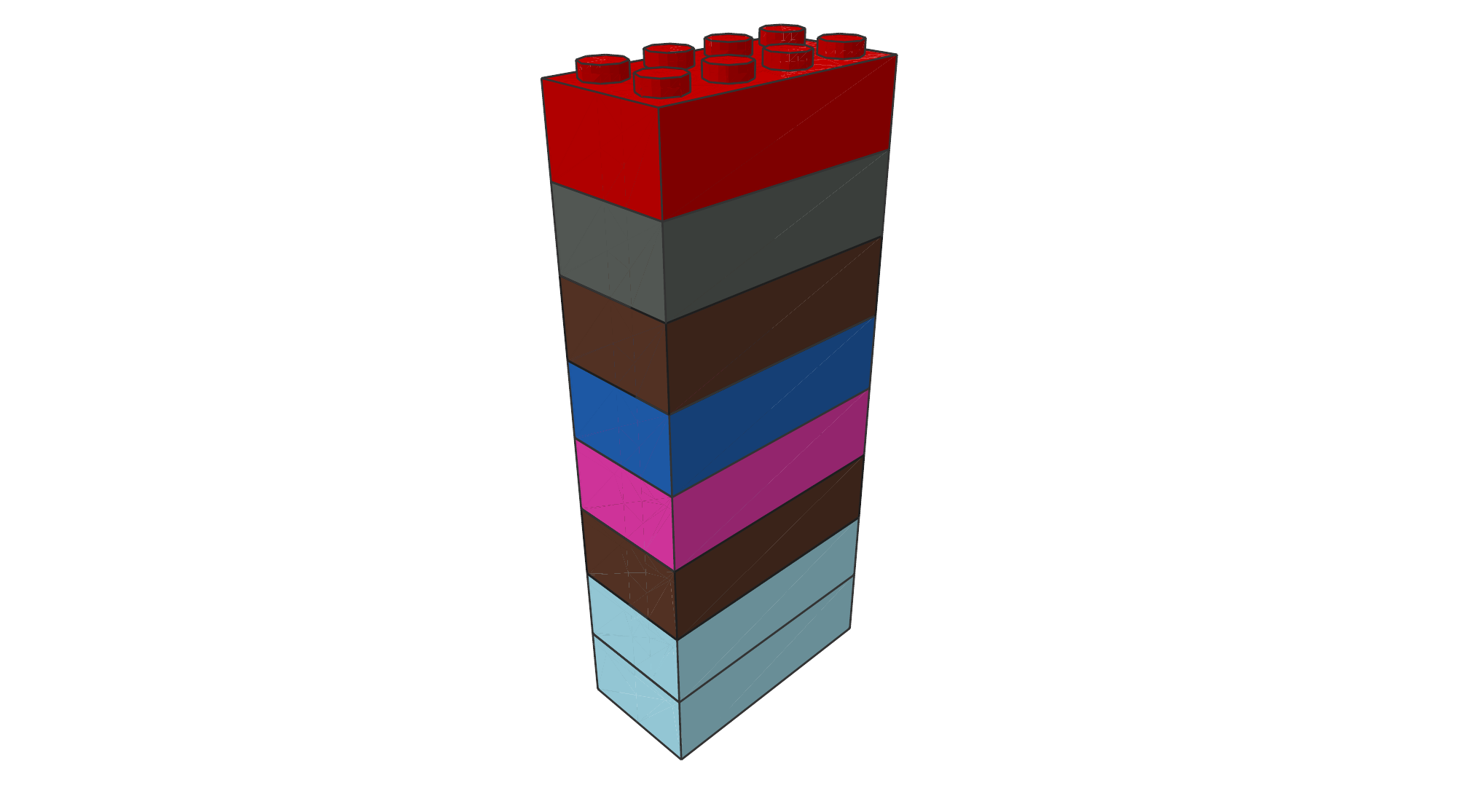}
        \caption{Bar}
    \end{subfigure} \hfill
    \begin{subfigure}[b]{0.3\textwidth}
        \includegraphics[width=\textwidth]{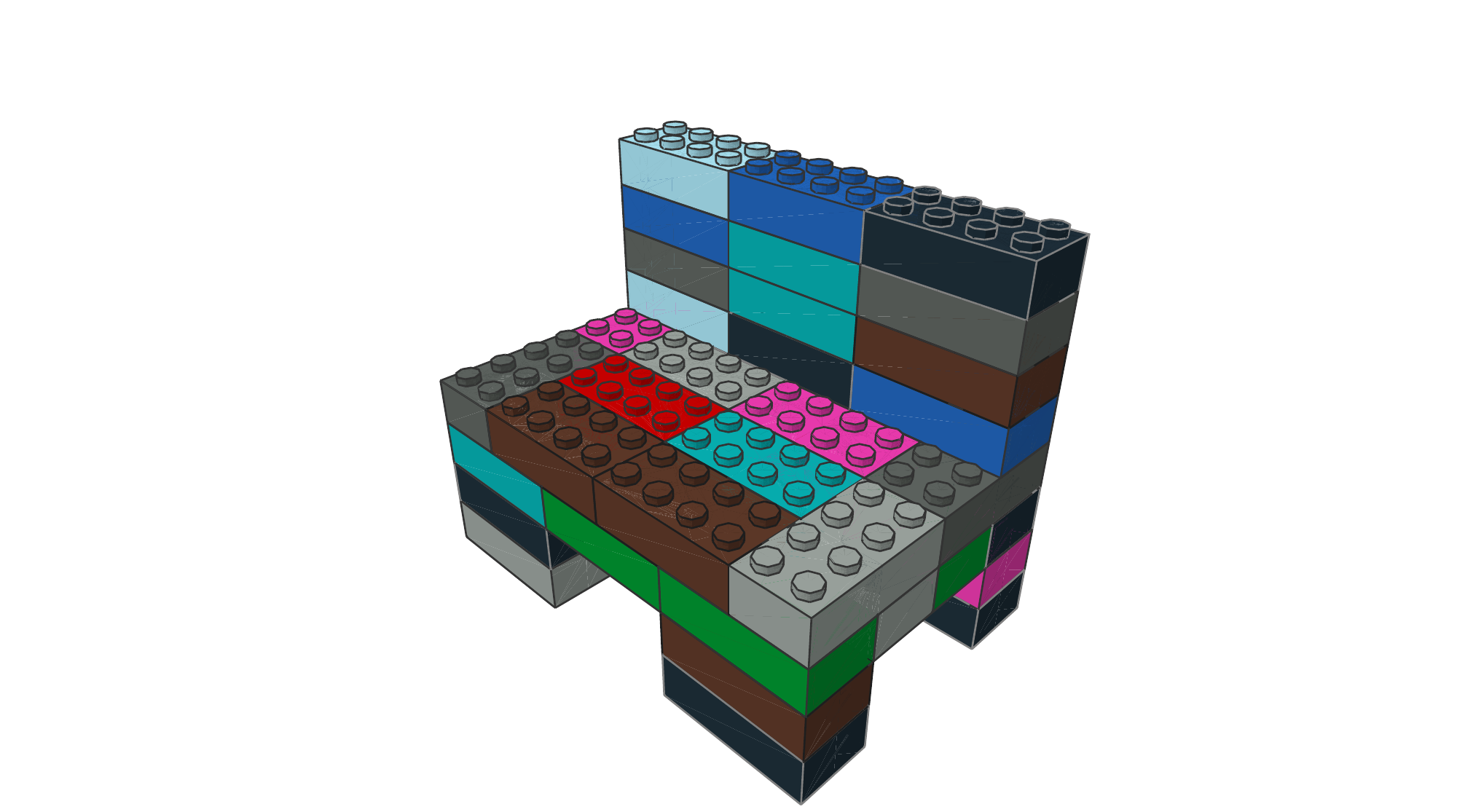}
        \caption{Bench}
    \end{subfigure} \hfill
    \begin{subfigure}[b]{0.3\textwidth}
        \includegraphics[width=\textwidth]{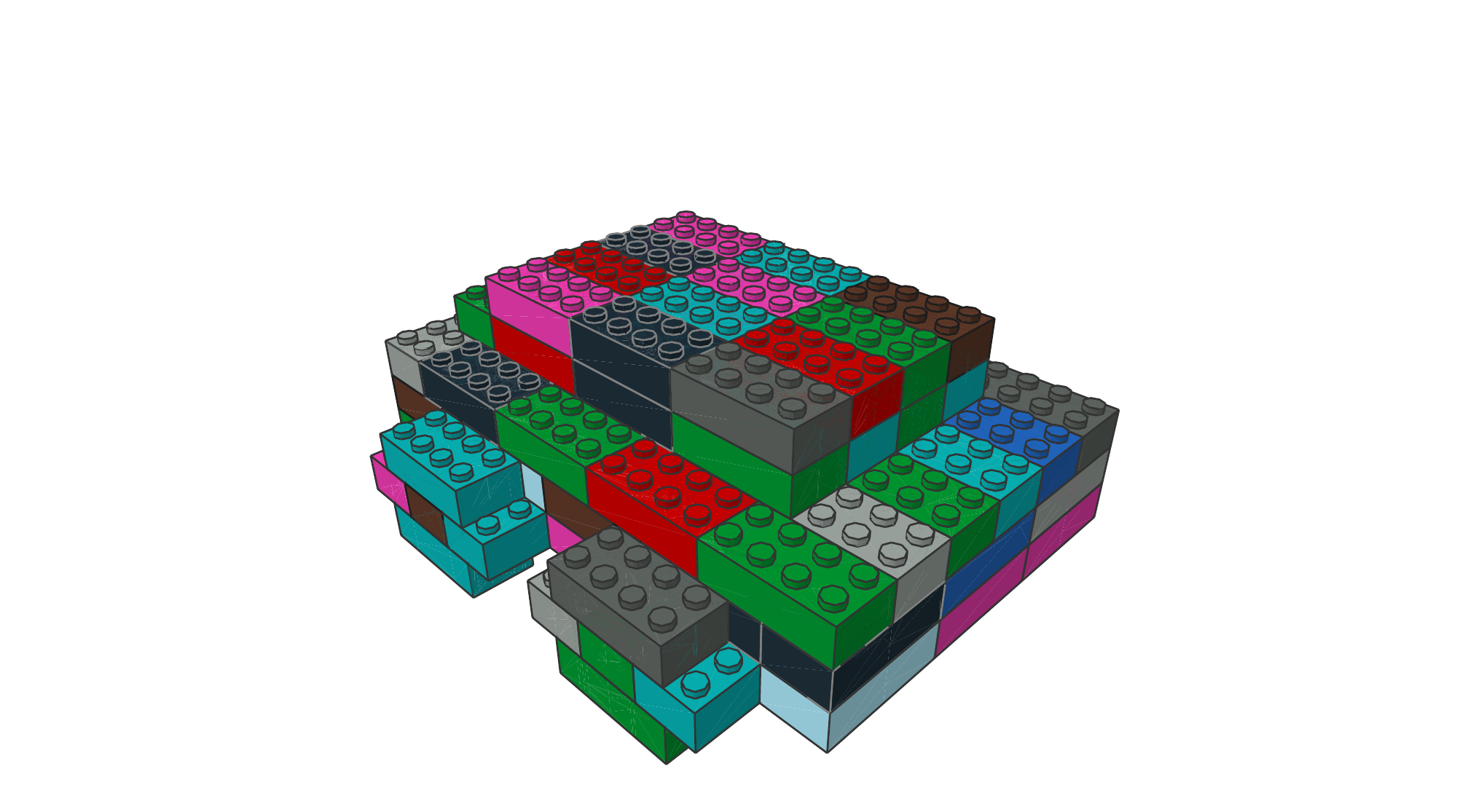}
        \caption{Car}
    \end{subfigure} \hfill
    \begin{subfigure}[b]{0.3\textwidth}
        \includegraphics[width=\textwidth]{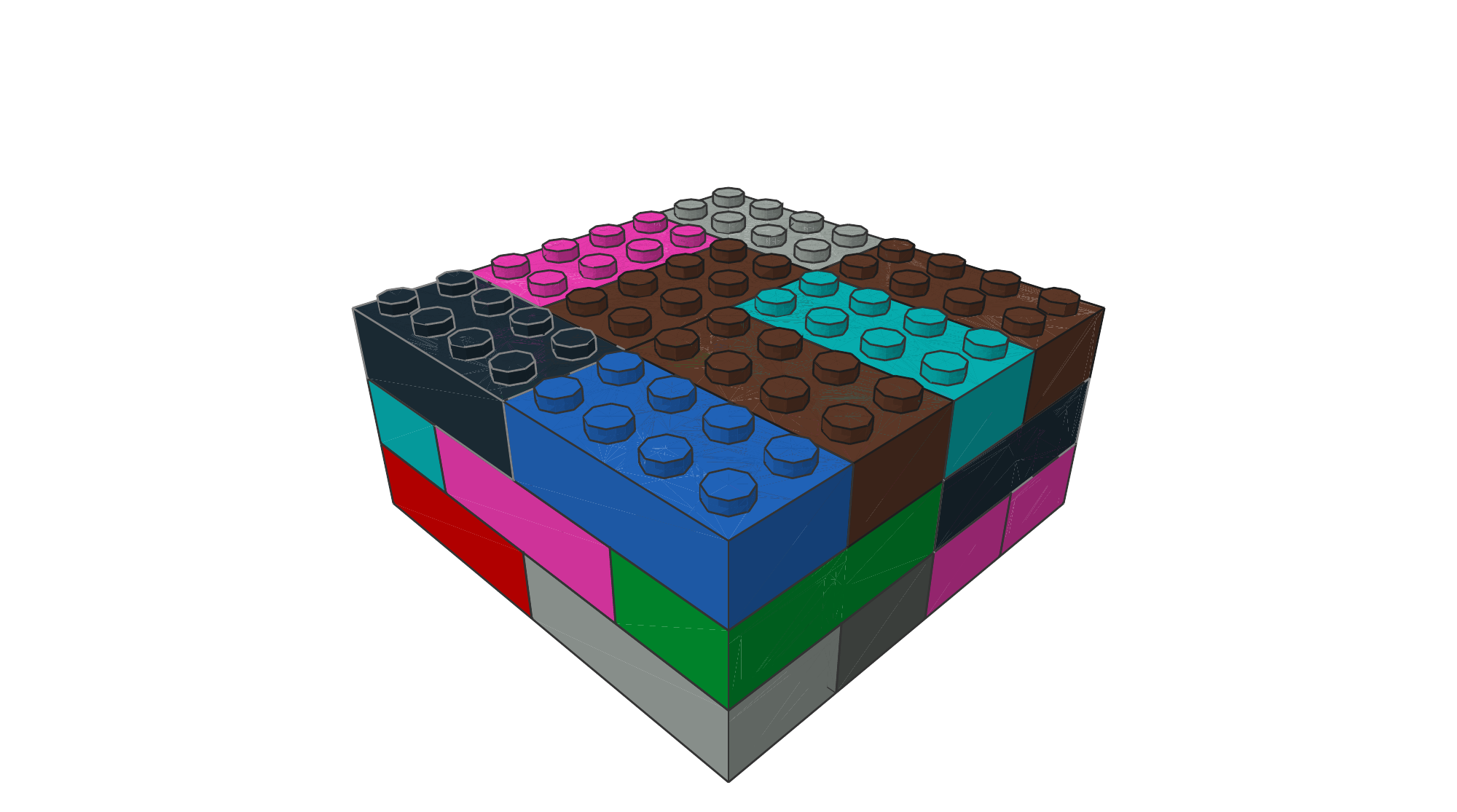}
        \caption{Cuboid}
    \end{subfigure} \hfill
    \begin{subfigure}[b]{0.3\textwidth}
        \includegraphics[width=\textwidth]{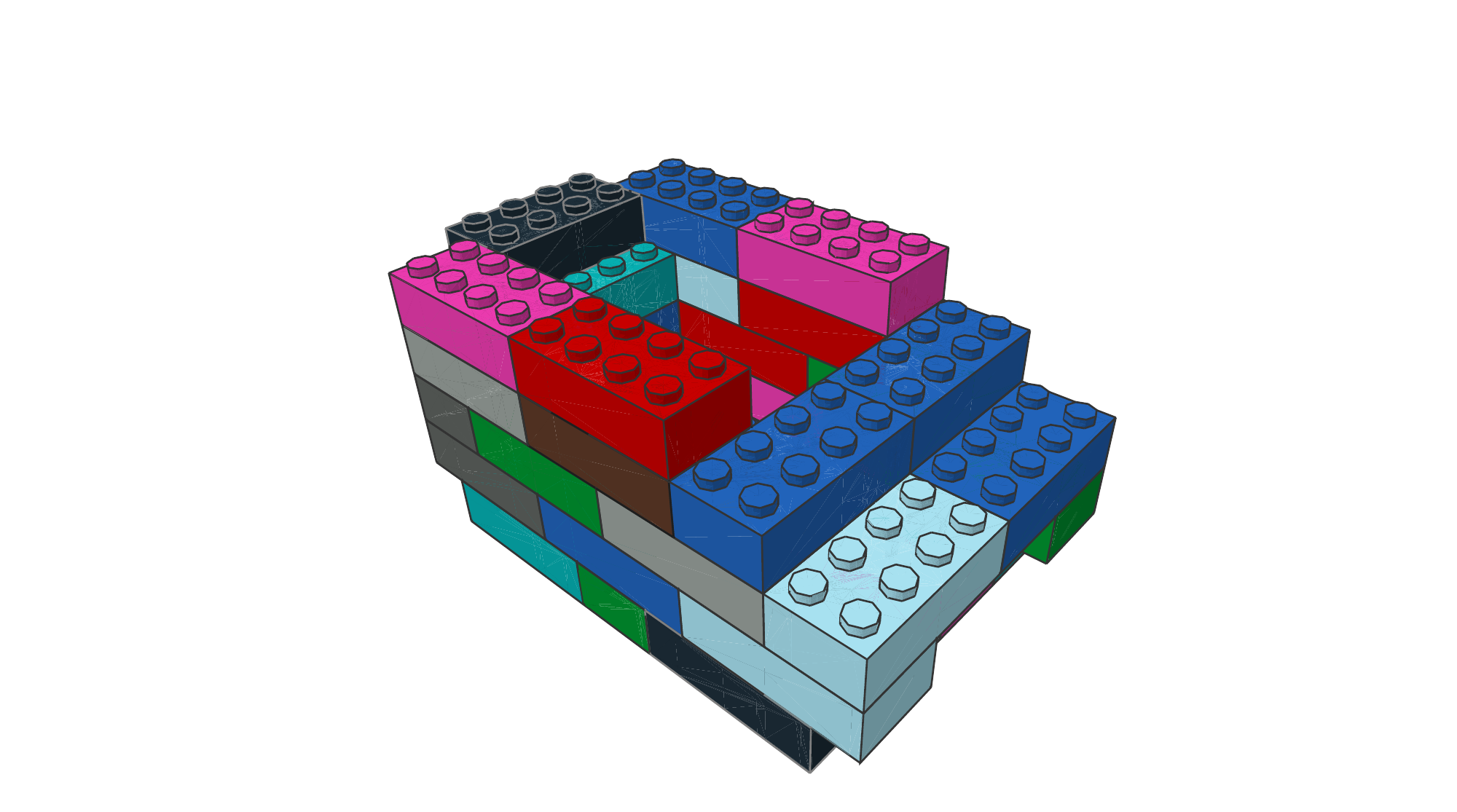}
        \caption{Cup}
    \end{subfigure} \hfill
    \begin{subfigure}[b]{0.3\textwidth}
        \includegraphics[width=\textwidth]{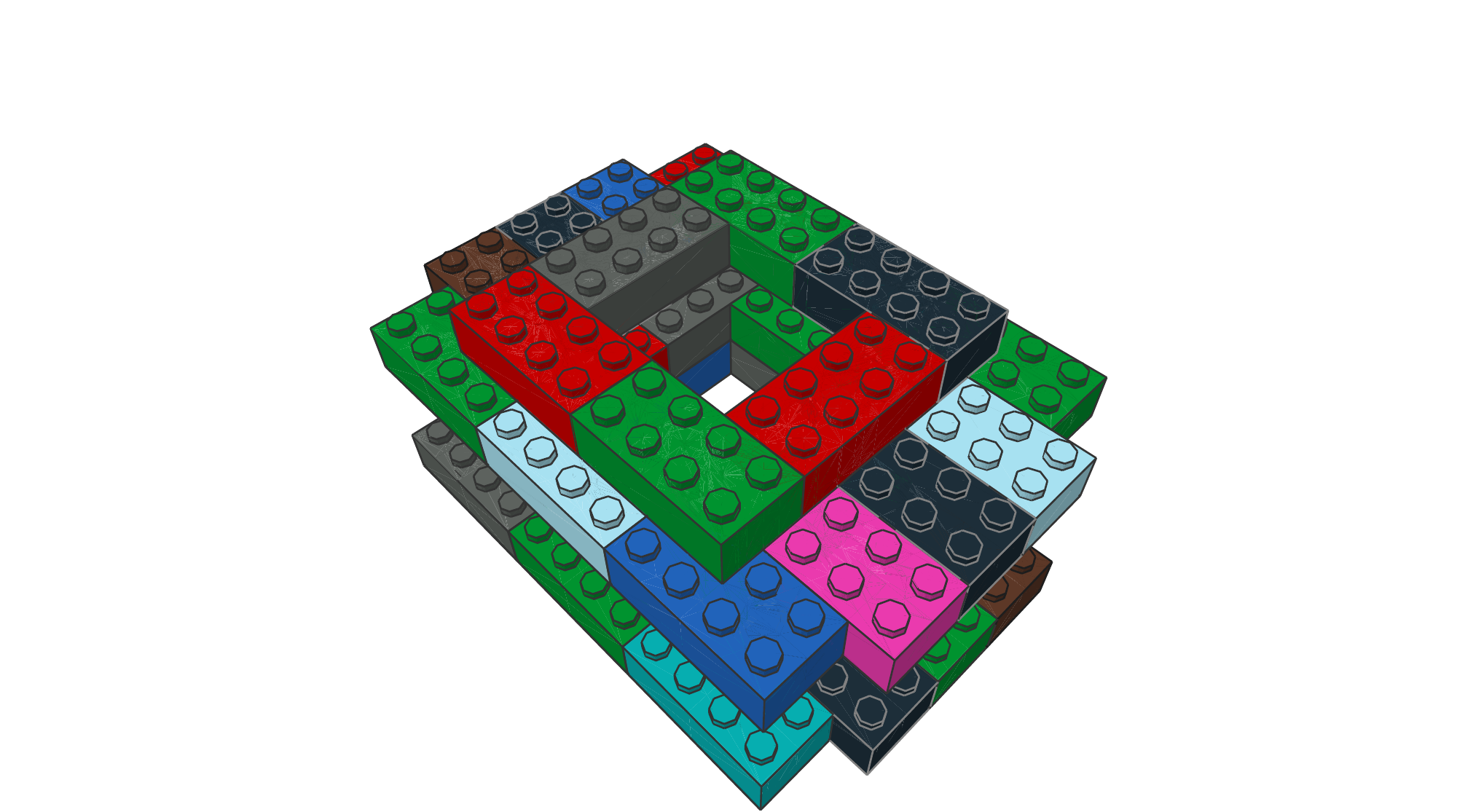}
        \caption{Hollow}
    \end{subfigure} \hfill
    \begin{subfigure}[b]{0.3\textwidth}
        \includegraphics[width=\textwidth]{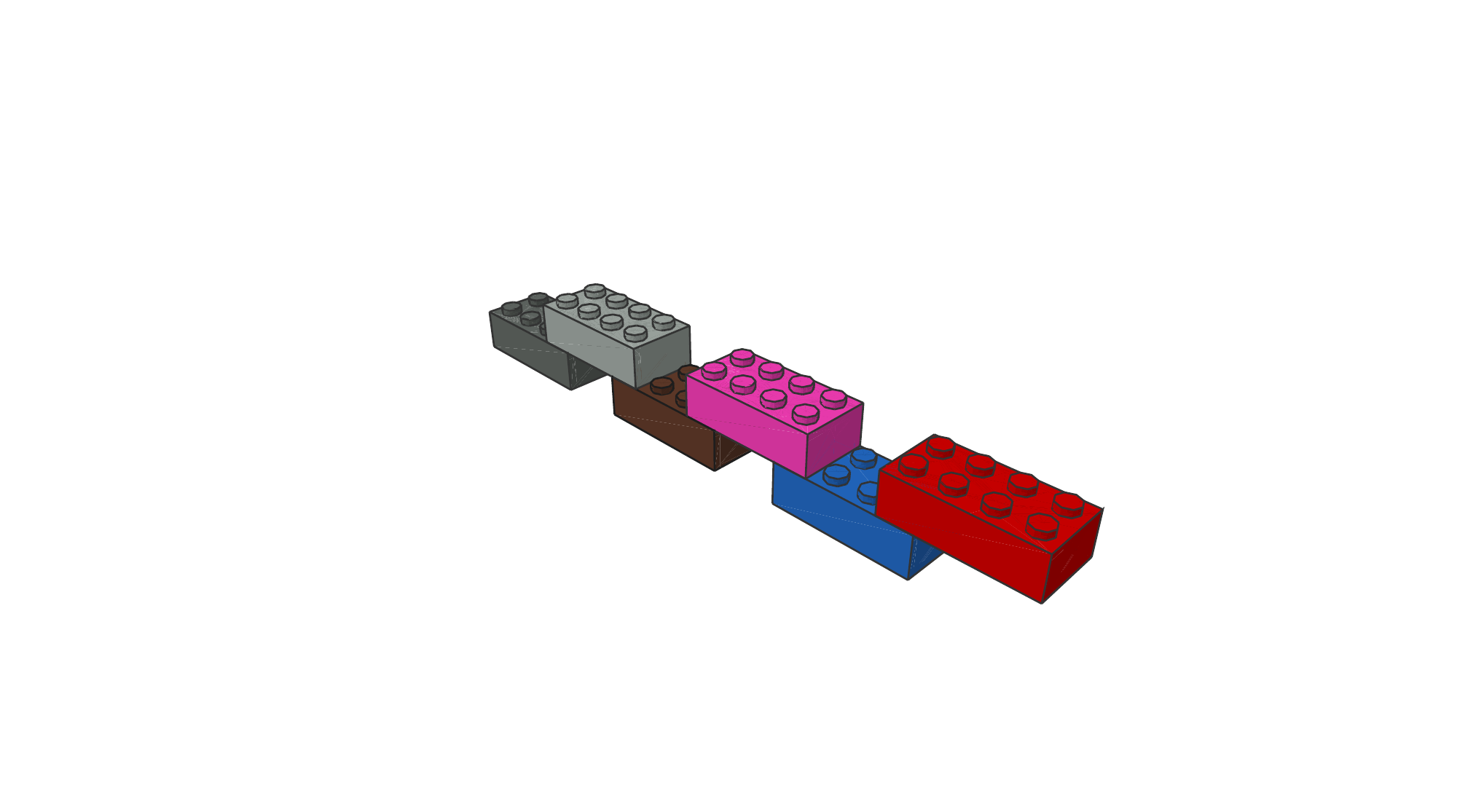}
        \caption{Line}
    \end{subfigure} \hfill
    \begin{subfigure}[b]{0.3\textwidth}
        \includegraphics[width=\textwidth]{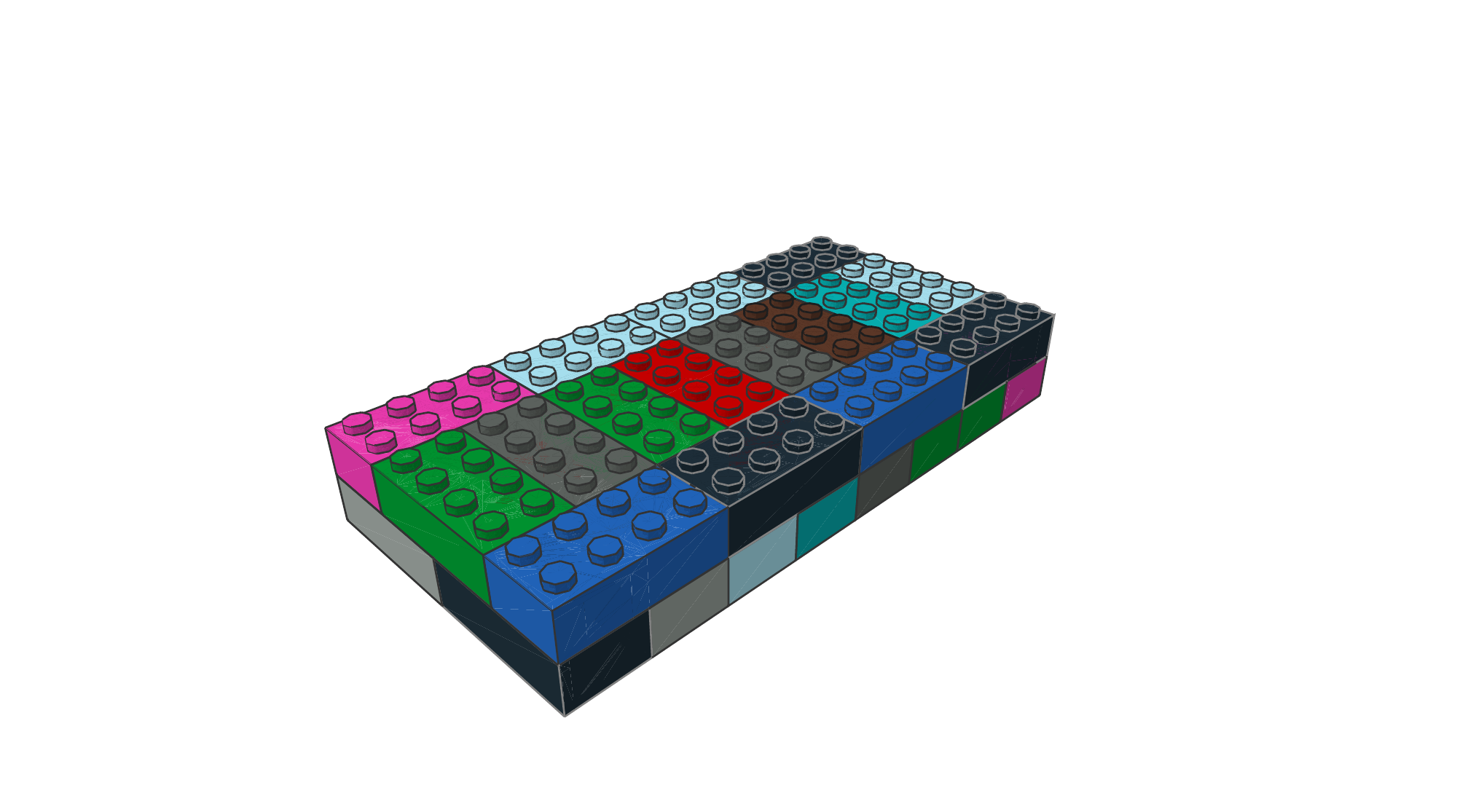}
        \caption{Plate}
    \end{subfigure} \hfill
    \begin{subfigure}[b]{0.3\textwidth}
        \includegraphics[width=\textwidth]{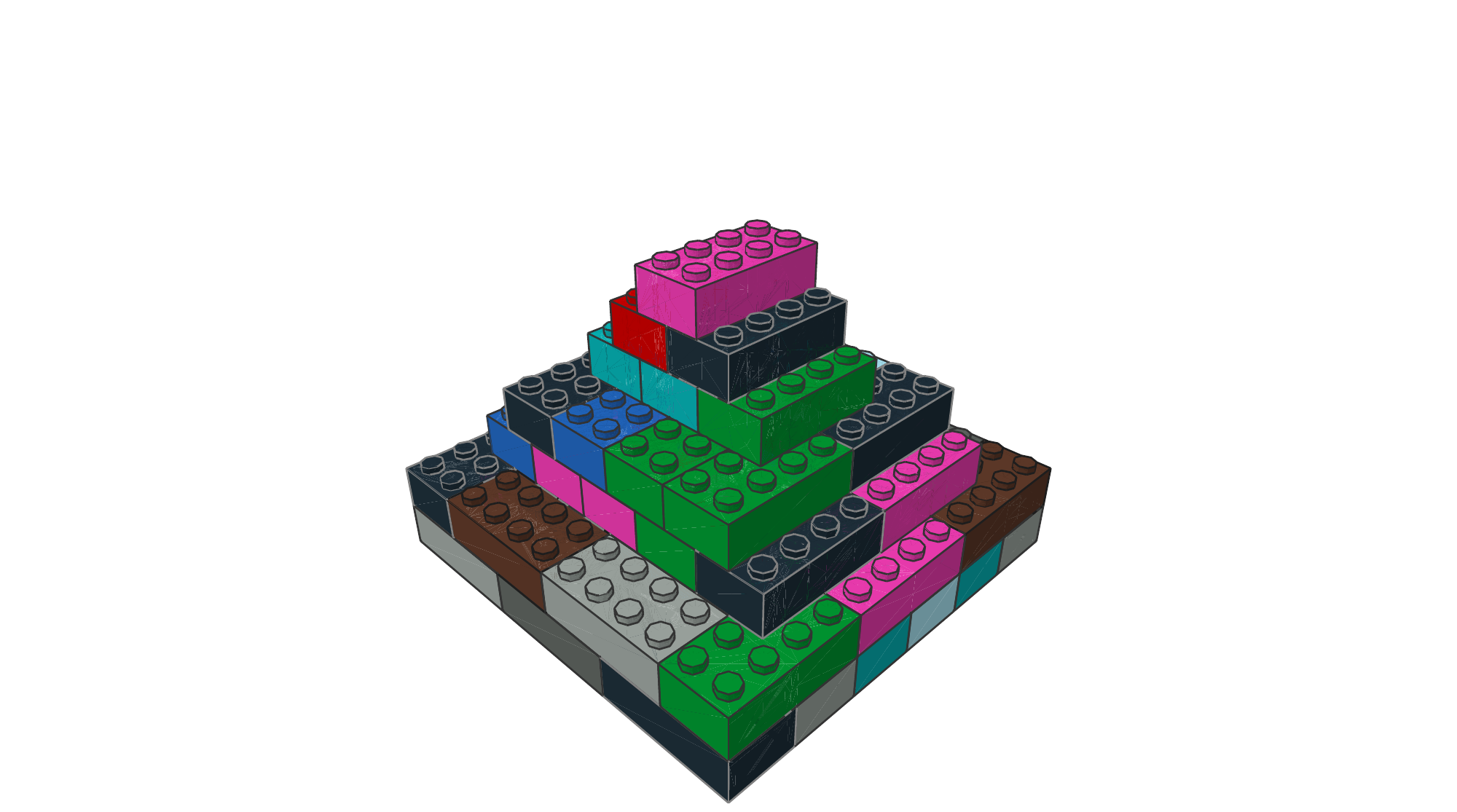}
        \caption{Pyramid}
    \end{subfigure} \hfill
    \begin{subfigure}[b]{0.3\textwidth}
        \includegraphics[width=\textwidth]{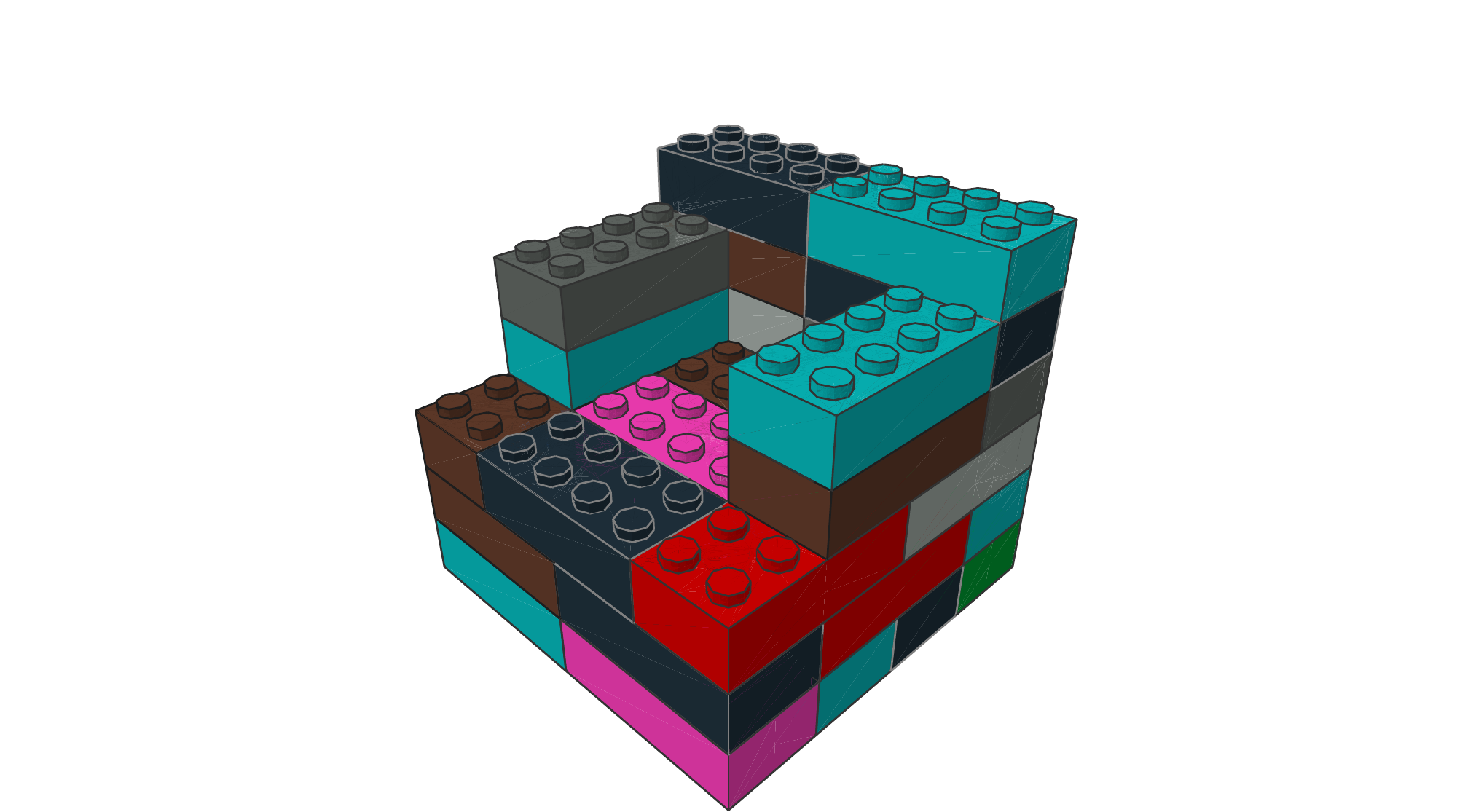}
        \caption{Sofa}
    \end{subfigure} \hfill
    \begin{subfigure}[b]{0.3\textwidth}
        \includegraphics[width=\textwidth]{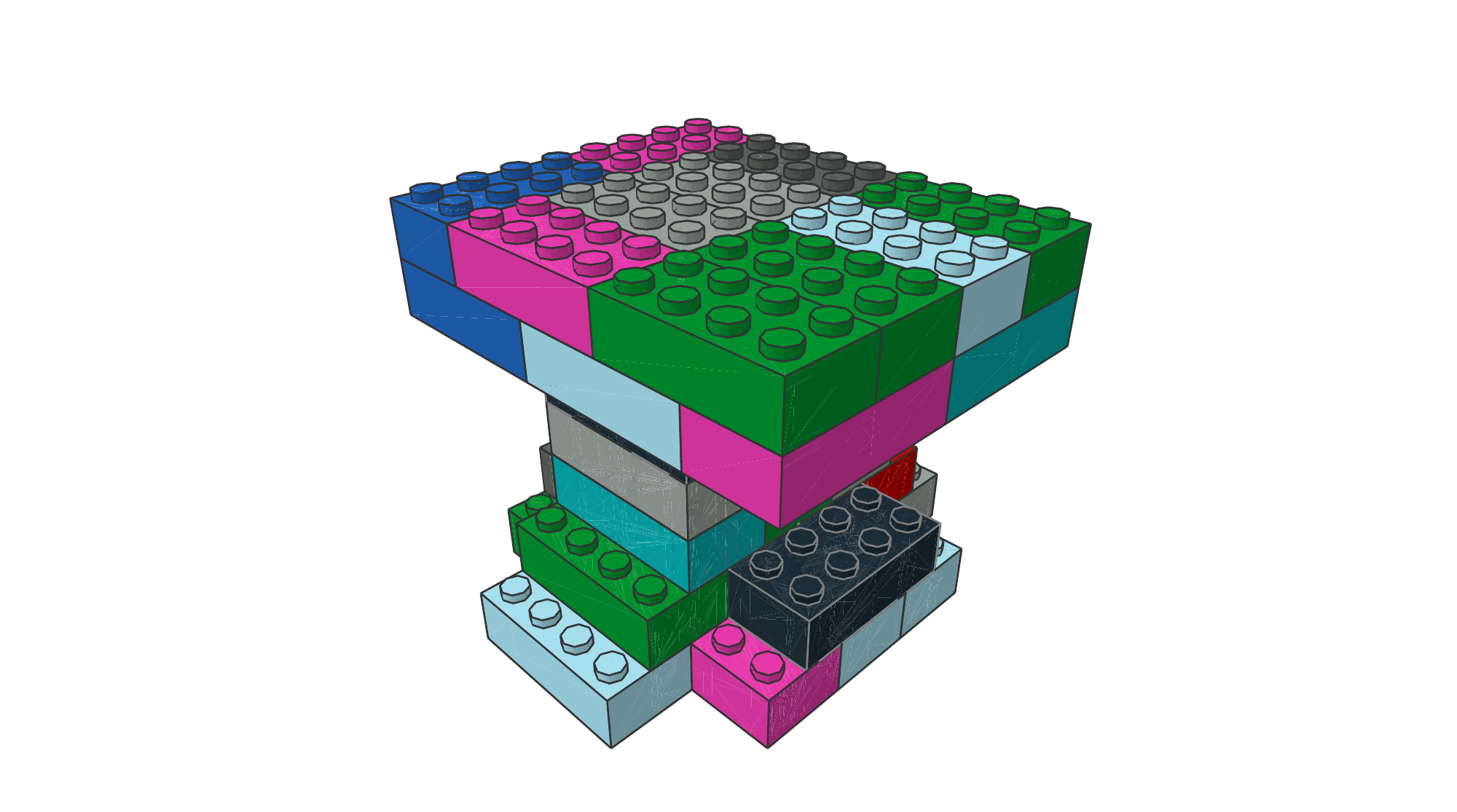}
        \caption{Table}
    \end{subfigure} \hfill
    \begin{subfigure}[b]{0.3\textwidth}
        \includegraphics[width=\textwidth]{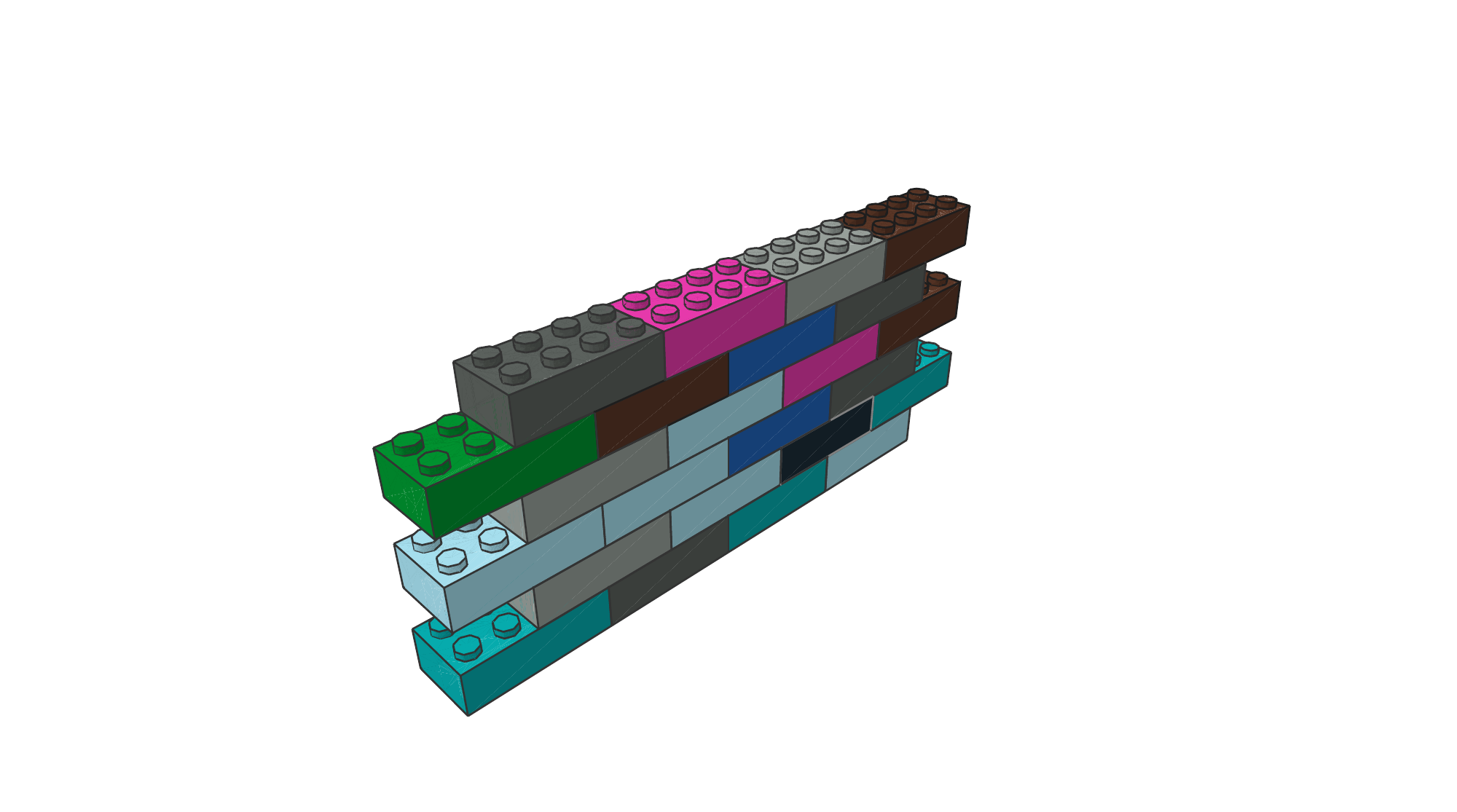}
        \caption{Wall}
    \end{subfigure} \hfill
    \caption{Representative examples of each class in the dataset}
    \label{fig:dataset-examples}
\end{figure}

The LEGO dataset consists of 12 classes and a total of 360 LEGO structures, each of which was created by one of twelve human subjects. Each class represents a relatively simple shape like ``bench'' or ``pyramid'', and a representative example of each class in the dataset is shown in Figure \ref{fig:dataset-examples}. The dataset has an average size of 56 bricks/nodes and 80 edges, and records the sequence of actions followed to create each structure; it is perfect for our needs. In addition, our LEGO graph representation is sensitive to rotational shifts, meaning the dataset can be readily augmented by rotating the original structures by 90, 180, and 270 degrees. We leverage this sensitivity to train generative models with a larger capacity than what would typically be possible with the original dataset alone. An example of each  class in the dataset is shown in Figure \ref{fig:dataset-examples}.

\end{document}